\documentclass[final,5p,times,twocolumn,authoryear]{elsarticle}

\usepackage{amssymb}
\usepackage{comment} 
\usepackage{booktabs}
\usepackage{lipsum}
\usepackage{xspace}
\usepackage{tipa}
\usepackage{float}
\usepackage{threeparttable}
\usepackage{adjustbox}
\usepackage{hyperref} 
\usepackage{amsmath}
\usepackage{float}
\usepackage{multicol}
\usepackage{url}
\usepackage{breakurl}

\begin{document}


\begin{frontmatter}



\title{Paved or unpaved? A Deep Learning derived Road Surface Global Dataset from Mapillary Street-View Imagery }

\author[first]{Sukanya Randhawa}
\ead{sukanya.randhawa@heigit.org}
\author[first]{Eren Aygün}
\author[second]{Guntaj Randhawa}
\author[first]{Benjamin Herfort}
\author[first,second,third]{Sven Lautenbach}
\author[first,second]{Alexander Zipf}

\affiliation[first]{organization={Heidelberg Institute of Geoinformation Technology (HeiGIT)},
            addressline={Berliner Str. 45 (Mathematikon)},
            city={Heidelberg},
             postcode={69120},
             state={Baden-Württemberg},
             country={Germany}}
\affiliation[second]{organization={GIScience Chair, Institute of Geography, Heidelberg University},
             addressline={Im Neuenheimer Feld 368},
             city={Heidelberg},
             postcode={69120},
             state={Baden-Württemberg},
             country={Germany}}

\affiliation[third]{organization={Centre for the Environment, Heidelberg University},
             addressline={Im Neuenheimer Feld 130.1},
             city={Heidelberg},
             postcode={69120},
             state={Baden-Württemberg},
             country={Germany}}

\begin{abstract}
We have released an open dataset with global coverage on road surface characteristics (paved or unpaved) derived utilizing 105 million images from the world’s largest crowdsourcing-based street view platform, Mapillary, leveraging  state-of-the-art geospatial AI methods. 
We propose a hybrid deep learning approach which combines SWIN-Transformer based road surface prediction and CLIP-and-DL segmentation based thresholding for filtering of bad quality images.
The road surface prediction results have been matched and integrated with OpenStreetMap (OSM) road geometries. 
This study provides global data insights derived from maps and statistics about spatial distribution of Mapillary coverage and road pavedness on a continent and countries scale, with rural and urban distinction.
This dataset expands the availability of global road surface information by over 3 million kilometers, now representing approximately 36\% of the total length of the global road network.
Most regions showed moderate to high paved road coverage (60-80\%), but significant gaps were noted in specific areas of Africa and Asia.
Urban areas tend to have near-complete paved coverage, while rural regions display more variability.
Model validation against OSM surface data achieved strong performance, with F1 scores for paved roads between 91-97\% across continents.
Taking forward the work of Mapillary and their contributors and enrichment of OSM road attributes, our work provides valuable insights for applications in urban planning, disaster routing, logistics optimization and addresses various Sustainable Development Goals (SDGS): especially SDGs 1 (No poverty), 3 (Good health and well-being), 8 (Decent work and economic growth), 9 (Industry, Innovation and Infrastructure), 11 (Sustainable cities and communities), 12 (Responsible consumption and production), and 13 (Climate action).

\end{abstract}




\end{frontmatter}


\section{Introduction}
\label{Introduction}

Road surface type information is a useful attribute for accurate travel time estimations, for disaster routing (humanitarian response) or for improved understanding of the drivers behind economic opportunity. It can furthermore facilitate data-driven urban planning, help to assess traffic related emissions or to understand biodiversity linkages and many such applications in environmental and climate science.

For instance, to efficiently plan suitable routes during disaster events, a routing service must take into account accurate estimated travel times by integrating critical data such as road type, surface conditions and road width from different sources and organizations.
Among other factors road surface type determines the vulnerability of a road network in the context of floods at the urban and national level highlighting the need of updated and complete data on road surface quality on a global scale \citep{agile-giss-2-4-2021, PAPILLOUD2020101548,petricola2022healthcare}.

Moreover, transportation plays a multifaceted role in the pursuit of (economic) development objectives and there is a strong association between per capita income and the magnitude and quality of road infrastructure and road conditions \citep{RePEc:wbk:wbrwps:921, SONG2021110538, PINATT2020100100}. According to the African Development Bank (AfDB), 53\% of roads in Africa are unpaved and are expected to remain so for the foreseeable future \citep{african2014tracking}.

OpenStreetMap (OSM) has been identified as as promising and growing open dataset to study road networks globally.
There have been studies on the assessment of completeness of OSM road data that have shown that globally, OSM is \(\sim \)83\% complete and \(\sim \)40\% of countries—including several in the developing world have a fully mapped street network \citep{worldusermap}.
Recent advancements in deep learning based approaches to derive geo-spatial information have improved the coverage of global scale open building and road datasets even further, providing data for previously unmapped or under-mapped regions \citep{facebook_map_with_ai, microsoft_building_footprints, google_open_buildings}.
However, it is important to go beyond detection of roads and buildings towards enrichment of (OSM) attributes of roads and buildings as it significantly enhances the utility of this data for various sectors, including transportation, urban planning, environmental studies, disaster management, and technology development.

Still, only few limited studies have reported OSM data quality beyond geometric completeness for certain regions \citep{su9060997, doi:10.1139/geomat-2021-0012}.
A detailed assessment of specific road attributes such as surface type is still missing. All we know is that the current state of road surface data in OSM is highly incomplete, where only 30-40\% of the global road network includes surface type data.
Nevertheless, in previous studies road surface condition prediction has been studied in the context of autonomous driving where different sensors (such as profilometers, inertial and acoustic sensors, smartphone motion sensors such as tri-axial accelerometers, gyroscopes, magnetometers and LIDAR) have been used for road defect detection and measurement \citep{ 8730820, s22249583, 9621967, 8476788}.
These approaches are both hard to reproduce and challenging to scale as such detailed sensor data is usually not openly available.

To address this overall data gap we investigate the potential of a novel source to derive road attribute information: Mapillary, a crowdsourcing platform to collect street-view imagery (SVI). 
Mapillary presents a very heterogeneous data source which contains imagery captured in a wide range of conditions and locations and is quite diverse in terms of scenes, weather or season \citep{Mapillarystate}.
Images are sourced from different devices (mobile phones, tablets, action cameras, professional capturing rigs) and differently experienced photographers.
Advantages of using SVI data include low-cost, rapid, high-resolution data capture and a unique pedestrian/vehicle perspective.
However, disadvantages include limited attribute information and unreliability for temporal analyses or availability of latest updated information \citep{biljecki_street_2021}. 

Rapid advances in areas of computer vision and deep learning algorithms have enabled researchers to detect, characterize and quantify different attributes of visual road-scene environments, thereby enhancing overall understanding of scene attributes (ranging from neighbourhood to building or street level attributes) such as safety \citep{LU201841, 10.1145/2733373.2806273}, greenness \citep{LI2015675} and population demographics \citep{Gebru_2017}.
Most of the times, element-level observation using object detection or segmentation models is performed to extract multiple category objects from street-view images \citep{ZHANG2018153}. Mapillary also provides the Vistas Dataset (large-scale street-level image dataset) containing 25,000 high-resolution images annotated into 66 object categories with additional, instance-specific labels for 37 classes. However, information about road surface is \textit{not} covered in the Vistas Dataset \citep{8237796}. 

As shown in some studies, it is also possible to extract high-dimensional visual features to enhance the understanding of images for evaluation of place distinctiveness and comparison of their similarities (w.r.t urban physical environment for example) by using deep convolutional neural networks (CNN) \citep{2019BuEnv.16006099W, Kang2020}.
Due to the explicit positional information encoding, vision transformers - utilizing the multi-head attention module \citep{NIPS2017_3f5ee243} - have shown superior performance in applications such as cross-view image geo-localization \citep{zhu2022transgeo}.
In particular, the SWIN transformer architecture has been successfully used for exhibiting generalized performance in various street-view image scenarios such as detection of urban physical disorder \citep{HU2023209}.
Since vision transformers learn spatially localized patterns, they make a good candidate for applications that involve understanding visual road scene environments such as road surface classification problems.

Our objective is to close the existing road attribute data gap, by releasing a first of its kind planet-scale dataset on road surface type (paved or unpaved) based on street-view imagery from Mapillary, in order to support humanitarian response, urban planning and progress towards the Sustainable Development Goals.
Building upon the success of deep learning based building and road geometry detection, we want to pave the way towards deep learning based enrichment of road attribute information.
We have made the dataset available at the following link TBD.
\textit{(Please note the data would be made available only after the acceptance of the paper)}
The overarching research aim of the paper could be broadly classified as follows: 

\begin{itemize}
    \item RQ1. To assess the reliability of computer vision based DL models for accurate prediction of road surface tags based on a street-view imagery from diverse global regions.
    \item RQ2. To derive insights based on integrating different geo-spatial data (urban, rural, human development index) to augment understanding of road surface condition across countries and regions.
\end{itemize}

In the next section, we outline the data preparation workflow. Section \ref{subsection:deep_learning_pipeline} provides details on the DL models employed. Next, we present the GIS analysis methods (Section \ref{gis_methodology}) and data insights and details for global road surface coverage (Section \ref{results}).

\section{Data Preparation}
\label{data_preparation}

\begin{figure*}[h]
  \centering
  \includegraphics[width=1.0\linewidth]{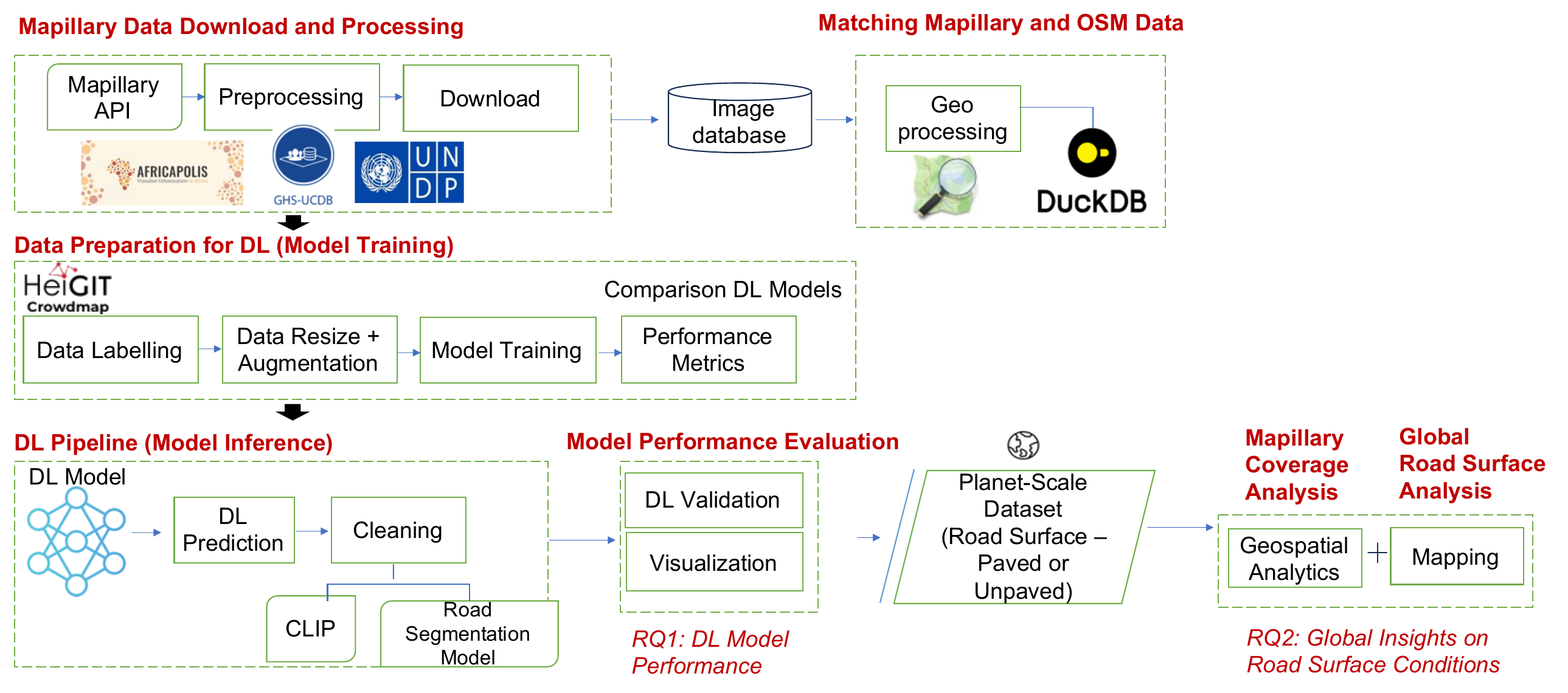} 
  \caption{Flow chart of the presented approach.}
  \label{fig:methodology}
\end{figure*}

The overall methodology is summarized in Figure \ref{fig:methodology}.
First, the Mapillary images were extracted and filtered for the entire globe (Section \ref{subsection:mapillary_data_download}).
Then, a balanced sub-dataset (in terms of class distribution) was created for training different deep learning models (Section \ref{subsection:deep_learning_preparation}).
The best performing model - the SWIN transformer - was deployed for generating a planet-scale dataset for road surface type information (Section \ref{subsection:deep_learning_pipeline}).

\subsection{Mapillary Data Download and Processing}
\label{subsection:mapillary_data_download}
As a first step, the Mapillary API tile endpoint (tiles.mapillary.com) was used to retrieve GPS sequences.
A sequence is a list of GPS coordinates where each node represents the location of an image. 
The requests for this endpoint were limited to 50,000 per day.
A tile map service (TMS) grid at zoom level 8 was created and intersected with continent polygons to filter out the ocean areas.
This amounted to a total of $16,423$ tiles.
Measured at the equator a single tile covers an area of roughly 150 km x 150 km.
All tiles containing Mapillary sequences were extracted (about $5,984$ tiles) as depicted in Figure \ref{mapillarycoveragemap}. 
A total of $11,031,802$ sequences were downloaded using the tiles endpoint.

Second, the Mapillary API graph endpoint (graph.mapillary.com/images) was used to get the metadata for multiple images with requests limited to 60,000 per minute.
This endpoint was used for download of all image points (with metadata including location and attribute data) corresponding to the sequences obtained in the previous step.
In total, over $11$ million sequence IDs have been processed which correspond to roughly $2,2$ billion image metadata points downloaded.
The attributes downloaded for each image have been listed in the Table 6 in Appendix. 
Figure \ref{fig:15} provides a closer look into the sequences and images on a city level from some key urban areas across the globe. 

 \begin{figure*}[h]
    \centering
    \includegraphics[width=1.0\linewidth]{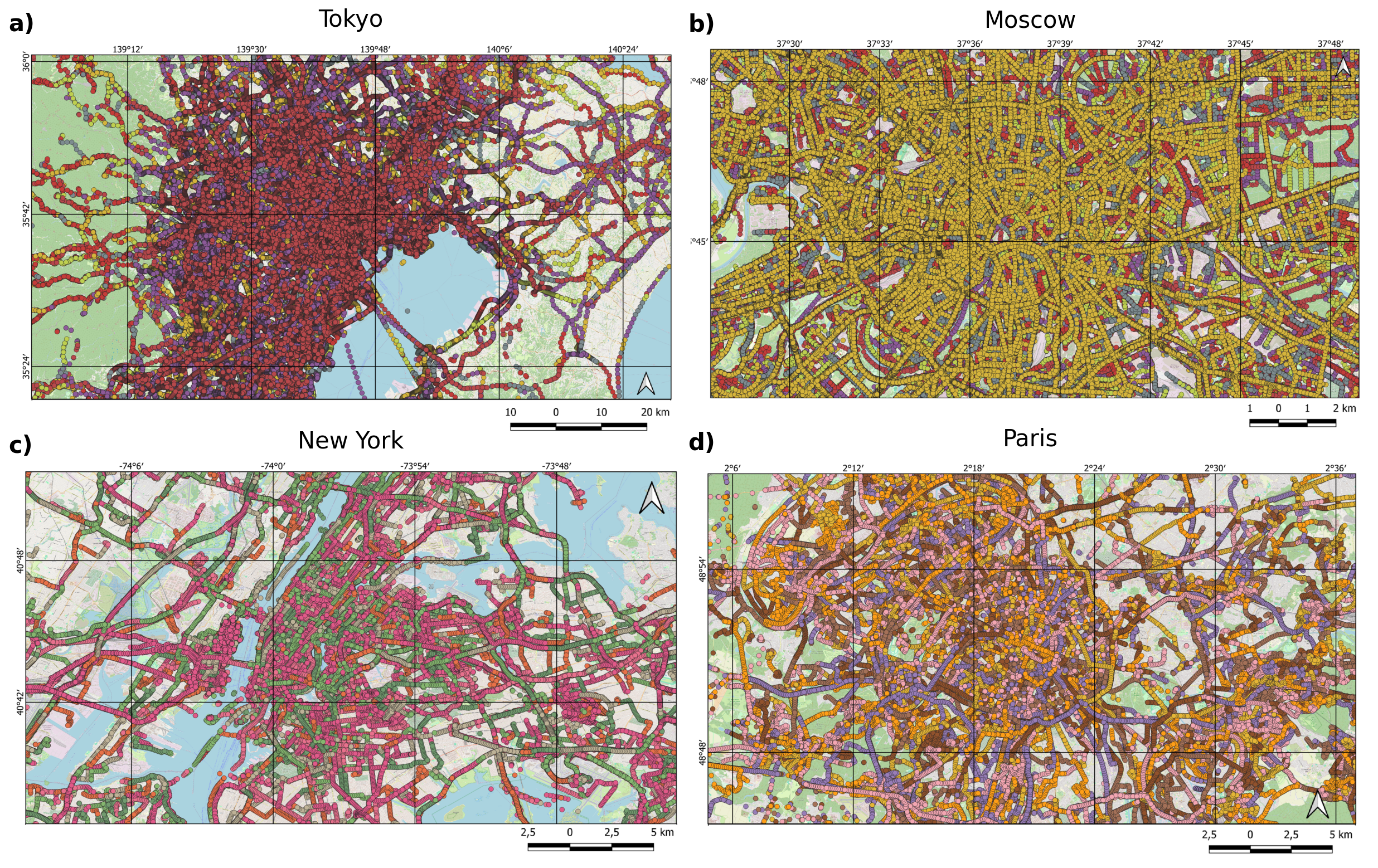}
    \caption{Map visualizations of road sequence data from various global urban areas. Each panel displays sequences that have been color-coded where possible, although the high volume of sequences in areas such as Tokyo and Moscow prevents distinct color coding. The different colors observed in sequences from San Francisco and Paris indicate various sequences; however, due to the limited color palette, a single color may represent multiple sequences.}
    \label{fig:15}
\end{figure*}

However, $373,889$ sequences out of roughly $11$ million could not be extracted because the download was blocked by the Mapillary API - $66\%$ of these sequences were located in Europe.
Figure 13 in Appendix provides a closer look into the distribution of the missing sequences.
Around $230,000$ out of $248,114$  missing sequences in Europe were located in Ukraine and were presumably blocked to prevent unintended usage for military action. 
Table \ref{tab:download_process} summarizes the results from the download process. 

Third, spatial filtering on the metadata was performed to select only the relevant images (preserving a spatial gap of 1000m for non-urban and 100m for urban areas) that would be later downloaded and fed into the Deep Learning models.
For data processing we used DuckDB, an open-source SQL database management system, which is optimized for complex query workloads and supports vectorized query execution, which significantly speeds up data processing \citep{10.14778/3554821.3554847}.
We used Apache parquet, a new, open source, column-oriented data file format designed for efficient data storage and retrieval.

Urban centers were identified by intersection with two datasets: 1) the Africapolis dataset \citep{africapolis2024} and 2) the Global Human Settlement Layer Urban Centres Database (GHS-UCBD) \citep{JRC115586} used for the rest of the world.
Next, each Mapillary sequence was thinned by selecting only one image along the sequence at 100m or 1000m depending on the type of area the sequence was mainly belonging to.
In addition, country information was added to these images based on a spatial join with country data from Overture Maps June 2024 beta edition \citep{overturemaps2024}.
Finally, we have added information about the Human Development Index (HDI) \citep{UNDPHDI} value per image.
This resulted in $104,523,781$ images globally.

\begin{table*}[h]
    \centering
    \caption{Download summary for the Mapillary sequences and images.}
    \begin{tabular}{lrrrrr}
        \toprule
        Continent & Sequences & Sequences & Sequences  & Images Metadata & Images Metadata  \\
             &  All & Missing (\#) & Missing (\%) & All & Filtered \\
        \midrule
        Africa          & 165,498     & 244       & 0.15          & 33,851,278    & 1,894,929    \\
        Asia            & 2,835,388   & 36,670    & 1.29          & 441,977,960   & 46,065,171    \\
        Europe          & 3,423,017   & 248,114   & 7.26          & 806,998,646   & 21,844,113    \\
        North America   & 3,186,923   & 80,292    & 2.52          & 711,519,573   & 26,129,136    \\
        Oceania         & 492,571     & 5,913     & 1.20          & 77,383,046    & 2,849,262    \\
        South America   & 928,405     & 3,656     & 0.39          & 138,317,309   & 5,741,170    \\
        \midrule
        Total           & 11,031,802  & 373,889   & 3.39          & 2,209,048,812 & 104,523,781  \\
        \bottomrule
    \end{tabular}
    \label{tab:download_process}
\end{table*}

\subsection{Data Preparation for Deep Learning}
\label{subsection:deep_learning_preparation}

In order to investigate if the road surface in the given street-view image was paved or unpaved, we treated this as an image classification problem.
We created a labeled dataset using the HeiGIT CrowdMap web application \citep{agile-giss-4-5-2023}.
During a mapathon at Heidelberg University in July 2023, 30 volunteer participants classified about 21,000 random Mapillary images from 39 countries in sub-Saharan Africa as either “paved”, “unpaved” or “bad imagery”.
The sub-Saharan Africa region ensured sufficient class distribution for both \textit{paved} and \textit{unpaved} categories, while the bad quality images were filtered out. These 21,000 images were split into training and validation sets (80:20).
To make sure that the prediction accuracy was not effected due to class imbalance, 8,500 images each of paved and unpaved images were used to create a balanced training dataset of a total of 17,000 images.

The images in the training set were size cropped from 2000x1200 pixels to 427x256 for optimizing time and space complexity.
This step helped in preserving the aspect ratio and maintaining a consistent input size.
The augmented images were enhanced further by applying the \textit{Random Flip}, \textit{RandAugment} \citep{cubuk2019randaugmentpracticalautomateddata} and \textit{Random Erasing} \citep{DBLP:journals/corr/abs-1708-04896} data augmentation methods.
Thereby, each image started off by being flipped horizontally, resulting in a mirror image, with a probability of 0.5, then underwent two randomly selected image augmentations out of fifteen available options, such as contrast adjustment, histogram equalization, color inversion, random rotation, increasing exposure, random horizontal (vertical) translation.
After that a Random Erasing was applied to the images, with a probability of 0.25. This results in generating images with various levels of occlusion as rectangular regions of the images are randomly selected and the pixels of this region are erased and replaced with random values.
Utilization of these data augmentation techniques allowed to increase the diversity of the images in the training set, thus effectively mitigating overfitting, improving the generalization ability and enhancing the model’s classification performance.

\subsection{Matching Mapillary and OpenStreetMap Data}

\begin{figure*}[h]
    \centering
    \includegraphics[width=1.0\linewidth]{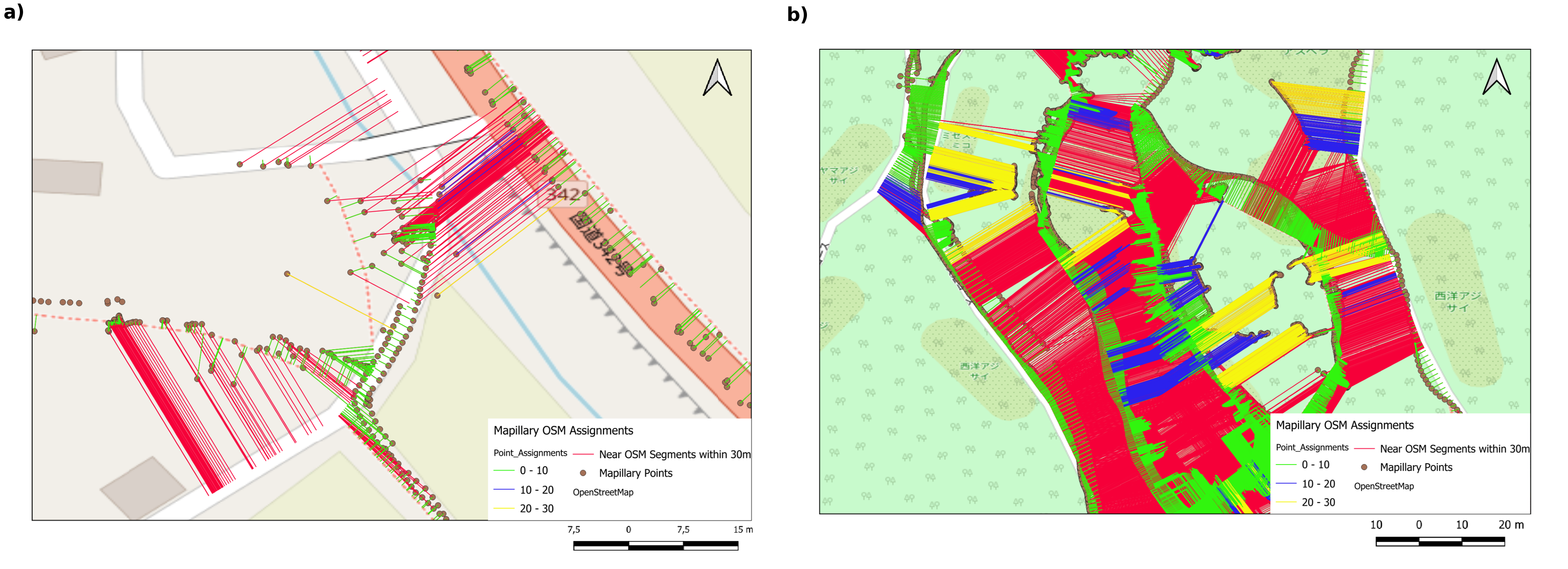}
    \caption{Comparison of Mapillary-OSM assignments from Sendai, Japan. Red lines correspond to the shortest lines to all OSM segments within 30 meters. The left sub-figure (a) shows examples were some image points are assigned to multiple OSM segments, particularly near intersections. The right sub-figure (b) shows examples for misallignments in non urban settings. OpenStreetMap is used as basemap for both sub-figures.}
    \label{fig:24}
\end{figure*}

Mapillary images provide a potential data source for adding OSM feature attribute information without the need to conduct a ﬁeld survey.
However, due to imprecise geo-referencing of images and OSM objects it was necessary to  map match the objects in the two data sets.
First, we extracted all relevant OSM road features via the ohsome API \citep{raifer2019oshdb} based on spatial location and selected all line geometries with the OSM tag key value 'highway'.
The matching algorithm was based on the shortest Euclidean distance between OSM road segment geometries and the Mapillary image coordinates (cf. Figure \ref{fig:24}).

We intersected the bounding boxes of the OSM road segments extended by a buffer of 30m with the Mapillary images.
All Mapillary image points within a 10 meter distance to OSM road segment were directly assigned to that road segment -  the threshold is based on the assumption of a GPS error tolerance of 5m.
Remaining points were assigned in a second and third step to the closest OSM segments within proximity of 20m and 30m respectively.
The results of the OSM assignment are depicted in Figure \ref{fig:24}.
It is also possible that some image points are assigned to multiple OSM segments, particularly near intersections as also depicted in Figure \ref{fig:24} (a).
Figure \ref{fig:24} (b) shows the matching in a forest/park area with sparse distribution of roads and highlights the misalignment that could possibly occur, underscoring the limitations of the algorithm in non-urban settings.
The advantage of this approach is that all the point outliers or bad points get filtered out automatically with no OSM segment assigned.
However, as OSM might not be completely mapped in the region, points, that correspond to road segments not mapped in OSM, are removed as well.

To standardize the distance-based assignment of each image point to the relevant OSM segment(s), we developed a normalized confidence index (Equation \ref{eq:12}).
This index captures the distance-based contrast  between the nearest OSM segment and other surrounding segments for each image point (in case of multiple segments within the buffered bounding box).
It becomes 0 for the nearest OSM segment as it is compared to itself and the index increases for segments that lie farther from the point.
As the \textbackslash{}textit\{percent\_diff\} approaches 1, the likelihood of a correct assignment to that OSM segment diminishes.
Higher values for both indicators imply greater uncertainty, suggesting that the image point might more accurately correspond to another (nearest) road segment.

\begin{align}
    \text{Index}_{\text{percent\_diff}} &= \frac{d_{{\text{current}}} - d_{\text{nearest}}}{d_{{\text{current}}} + d_{\text{nearest}}}
    \label{eq:12}
\end{align}

As a final step towards data enrichment all the matched Mapillary image points and their corresponding label (paved or unpaved) were aggregated per OSM road segment ID.
This was done via a distance-based weighted average approach wherein the closer points matched to a segment were given higher weights than the ones further away.
Based on a this weighted average score, the road attribute paved or unpaved was determined and assigned per OSM road segment.

\section{Deep Learning Pipeline}
\label{subsection:deep_learning_pipeline}

We trained a classical convolutional neural network (CNN) deep learning model (ResNet) \citep{he2016deep} as well as two recent model architectures: a SWIN Transformer model \citep{Liu_2021_ICCV} and ConvNext \citep{liu2022convnet} as a CNN that borrows elements from Vision Transformer models.
We used the models from OpenMMLab (https://openmmlab.com/) which is an open-source algorithm platform that is based on PyTorch deep learning framework for model training and evaluation.
As expected the performance of ConvNext and SWIN Transformer models is slightly better than ResNet (cf. Table \ref{tab:comparison-results}), although all three models performed quite well for this problem with macro average F-1 scores greater than 98\%, while the training dataset was comprised of images from the African continent.

The different model architectures process their inputs very differently. ResNet allows intricate feature extraction due to the incorporation of residual or skip connections which leads to increased model robustness. However, ResNet relies on predefined patterns and hierarchical feature extraction. In comparison, the Vision Transformer models are based on a self-attention mechanism. This mechanism allows the model to focus more on an understanding of the relationships between different elements within an image and enables thereby the model to interpret images as a whole. This is based on the use of non-overlapping patches rather than relying on stacked convolutional layers. Hierarchical Transformer models (e.g. SWIN Transformers as used here) have been shown to outperform normal Vision Transformer models on a wide variety of vision tasks \citep{Liu_2021_ICCV}.
On the other hand, ConvNext, which is a pure convolutional model, was inspired by the design of Vision Transformers. The ConvNeXt architecture used a pyramidial structure and achieved competitive performance on various vision tasks. ConvNext and SWIN Transformer are both equipped with similar inductive biases, but differ significantly in the training procedure and macro/micro-level architecture design.

\begin{table}
  \caption{Comparison ResNet, ConvNext and SWIN Transformer for road surface prediction}
  \label{tab:comparison-results}
  \begin{tabular}{ccccccl}
    \toprule
     \hline
    Model & Precision & Recall & F-1 Score\\
     \hline
    \midrule
    ResNet & 98.11 & 98.11 & 98.11\\
    ConvNext & 98.63 & 98.63 & 98.63 \\
    SWIN-Transformer & 98.68 & 98.68 & 98.68\\
  \bottomrule
\end{tabular}
\end{table}

Even though all the models showed robust performance, we used a SWIN-Transformer (Hierarchical Vision Transformer using shifted Windows) image classification model which had been pre-trained on the Imagenet-1k dataset \citep{deng2009imagenet} and fine-tuned on our self-annotated street view imagery dataset for the planet-scale application.
Our decision is based on the knowledge that the model architecture is known to establish long-range dependencies \citep{Liu_2021_ICCV}, effectively extracting global features from the images.
Given the heterogeneity of the Mapillary street-view imagery across different global regions, this model architecture was assumed to offer a higher potential for accurate road surface classification based on the road surroundings.
With label smoothed cross entropy loss, the model was trained for 120 epochs, employing the Adam optimizer with weight decay.


As is expected in a binary classification problem, all the images were classified as having paved or unpaved roads. 
However, some anomalous images that did not contain any street were nevertheless classified by the model with respect to road surface information.
These bad images for example contain only side-views from a street containing partial views to sidewalks or a track.
To circumvent the problem of images with no or partial roads, we used a combination filter with the following two models:

\begin{enumerate}
    \item A fast semantic segmentation model \citep{xu2023pidnetrealtimesemanticsegmentation}, pre-trained on the "cityscapes" dataset, was used to filter out images with less than 10\% road pixels.
    \item A CLIP (Contrastive Language–Image Pre-training) model based on the ViT-L/14 Transformer architecture \citep{radford2021learningtransferablevisualmodels} as a zero shot classification model was used to filter out images without roads. Two classes i.e \textit{a photo of a road} and \textit{a photo with no road in it} were defined (based on a probability threshold) and used for removal process. 
\end{enumerate}

\begin{figure}[htbp]
  \centering
  \includegraphics[width=\linewidth]{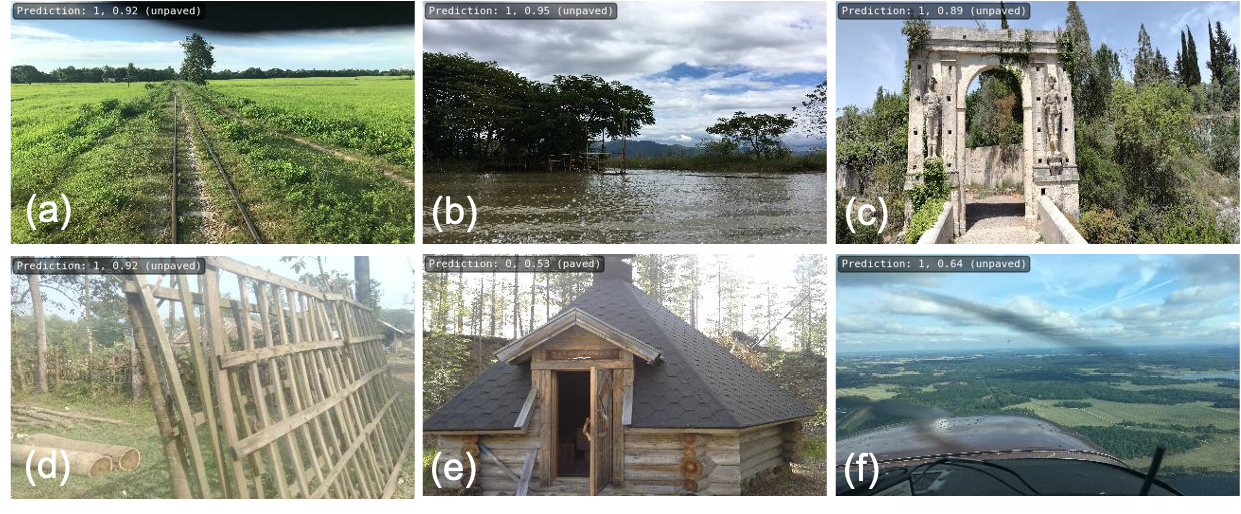}
  \caption{Examples of filtered images that did not contain any road using the CLIP model. }
  \label{remimg}
\end{figure}

The final combination filter removes images where the road pixel coverage (in terms of proportion of total image pixels) is less than 10\% and the CLIP model predicts that the image is “a photo with no road in it” (for any probability score) or where the clip model says that the image is “a photo with no road in it” with a probability greater than 90\%.
In total, these comprised approximately 4-5\% of the total images.
However, a few images with no roads might have escaped our cleaning procedure.
Fig.\ref{remimg} (a)-(f) shows examples of such removed images collected from different global regions.
These mostly include images of some random house, railroads, fences, buildings or then aerial imagery.

\section{Data Analysis}
\label{gis_methodology}

\subsection{Model Evaluation Methods}
To investigate model performance, we first assess the reliability of our model designed to predict road surface using street-view images. To ensure the model’s accuracy, we employed Grad-CAM (Gradient-weighted Class Activation Mapping) for visual explanations \citep{Selvaraju_2019}. This method allowed us to inspect how well the SWIN Transformer model identifies paved and unpaved roads.

Furthermore, we compare our model results with OSM data in order to understand to what extent our predicted road surface corresponds to the already mapped OSM road surface tags.
To assess the accuracy a confusion matrix for each continent was derived based on the predictions, depicting the True and False Positives and Negatives and correspondingly calculated measures thereof -  Accuracy, F-1 score, Precision and Recall.
True Positives are images where model prediction and OSM agree on the existence of a paved road, whilst True Negatives are images where both datasets agree that the road is unpaved.
Consequently, False Positives are images where the model predicted a paved road, but OSM indicated an unpaved road.
False Negatives are the cases where the model prediction resulted in an unpaved road, whereas OSM labelled as a paved road.
We evaluated the confusion matrix and derived performance scores for each continent.
The OSM road attributes were simplified as "paved" or "unpaved" based the following OSM tag values: 

\begin{itemize}
    \item Paved: surface in (paved, asphalt, chipseal, concrete, concrete:lanes, concrete:plates, paving\_stones, sett, unhewn\_cobblestone, cobblestone, bricks, metal, wood)
    \item Unpaved: surface in (unpaved, compacted, fine\_gravel, gravel, shells, rock, pebblestone, ground, dirt, earth, grass, grass\_paver, metal\_grid, mud, sand, woodchips, snow, ice, salt)
\end{itemize}

\subsection{Global Road Surface Analysis}
We analyzed how the Mapillary coverage of OSM roads varies across space.
To assess this spatial heterogeneity, we calculated the ratio of total Mapillary sequence length vis-a-vis total OSM segment length for each tile of a zoom level 8 map of the whole world.
As a single OSM road segment can be (partially) covered by several overlapping Mapillary sequences, we calculated the total Mapillary coverage length per tile as the sum of the longest Mapillary sequence for each OSM segment in that tile.
In our calculations we further distinguished Mapillary coverage between urban and rural area and for various road type classes as indicated by OSM highway tag values.
 
Second, we investigated the surface pavedness ratio for all roads which were covered by Mapillary data on a global scale using a zoom level 8 tiles map.
Again, this information has been disaggregated using urban / rural images and per road type class.
Additionally, we derived box plots to inspect the distribution of surface pavedness across continents and road type. 
Finally, we run a regression analysis to investigate the correlation between average road pavedness and HDI values per country.

\section{Results}
\label{results}

The upcoming sections presents our findings, where we evaluate the performance of our deep learning models on a global scale by comparing correct and incorrect predictions with OSM data via confusion matrix and standard machine learning performance metrics.
 Moreover, we analyse the spatial distribution of Mapillary data coverage along with OSM road surface information, delve deeper into the paved ratio analysis to understand the emerging global patterns for road surface infrastructure on continent and country level. 

\subsection{Model Performance Evaluation}

\begin{figure}[h]
  \centering
  \includegraphics[width=\linewidth]{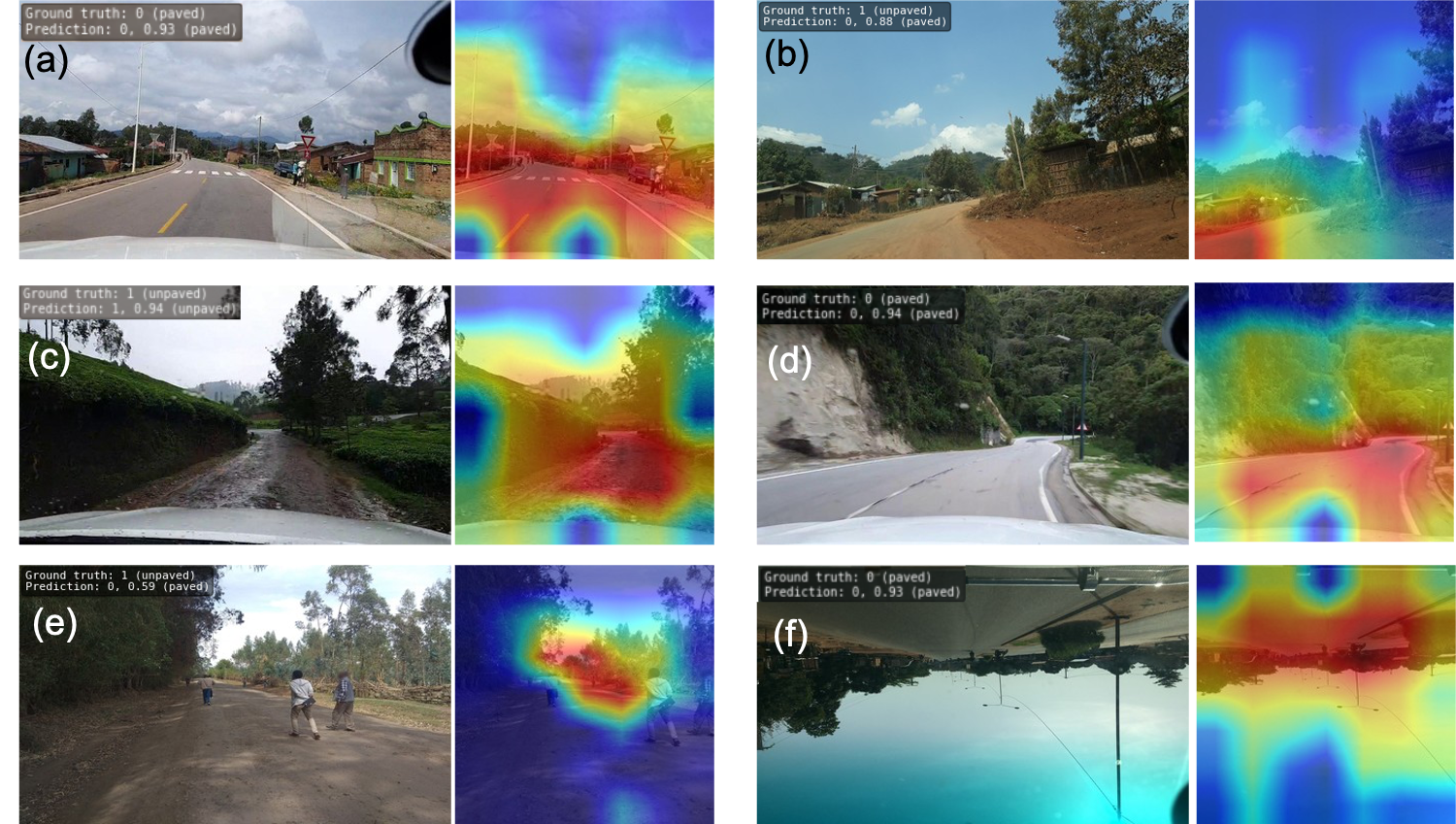}
  \caption{Visualization of road surface classification. Warmer colors (red, orange, and yellow) denote higher intensity activation levels, whereas cooler colors (e.g., blue and green) denote lower activation levels.}
  \label{actmaps}
\end{figure}

By reviewing the activation maps (cf. Fig. \ref{actmaps}), we observed that the model could accurately highlight areas corresponding to road surfaces, with warmer colors denoting areas of higher activation, indicating that these regions contributed significantly to the surface classification.
Interestingly, the model maintained robustness even when facing complex or ambiguous images, such as those with partial road visibility or covered with dust and sand (Fig. \ref{actmaps}(b), (e), and (f)).
Despite the variations in image quality, lighting, and scenery, the model's predictions remained consistent, providing confidence in its generalization capabilities across diverse conditions.

\begin{figure}[h]
  \centering
  \includegraphics[width=\linewidth]{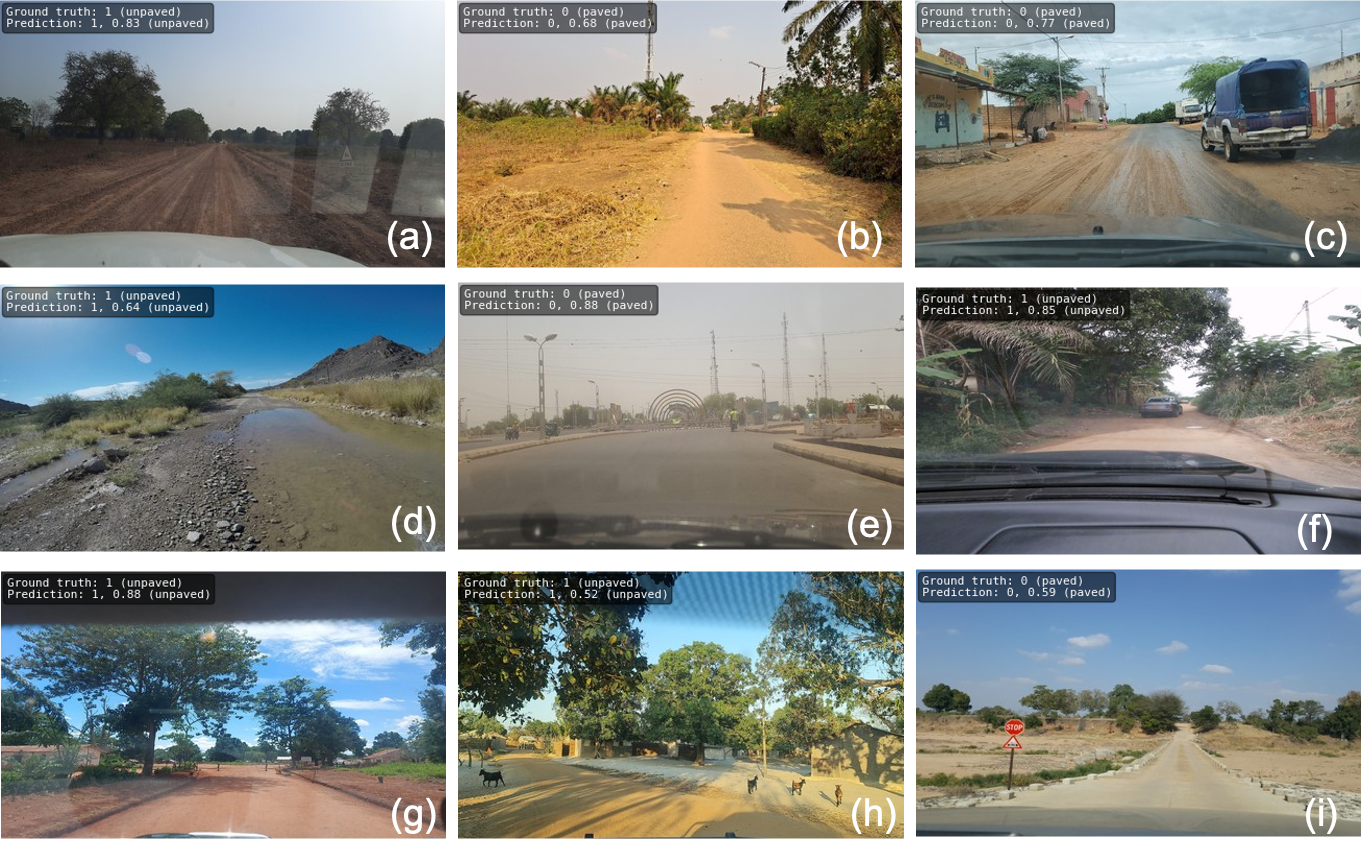}
  \caption{ Examples of True Positives  and True Negatives for the SWIN Transformer based classification model for road surface prediction on Mapillary based street-view images.  Ground Truth, Prediction and confidence score are indicated in the top left corner of each image
  }
  \label{tps}
\end{figure}

\begin{figure}[h]
  \centering
  \includegraphics[width=\linewidth]{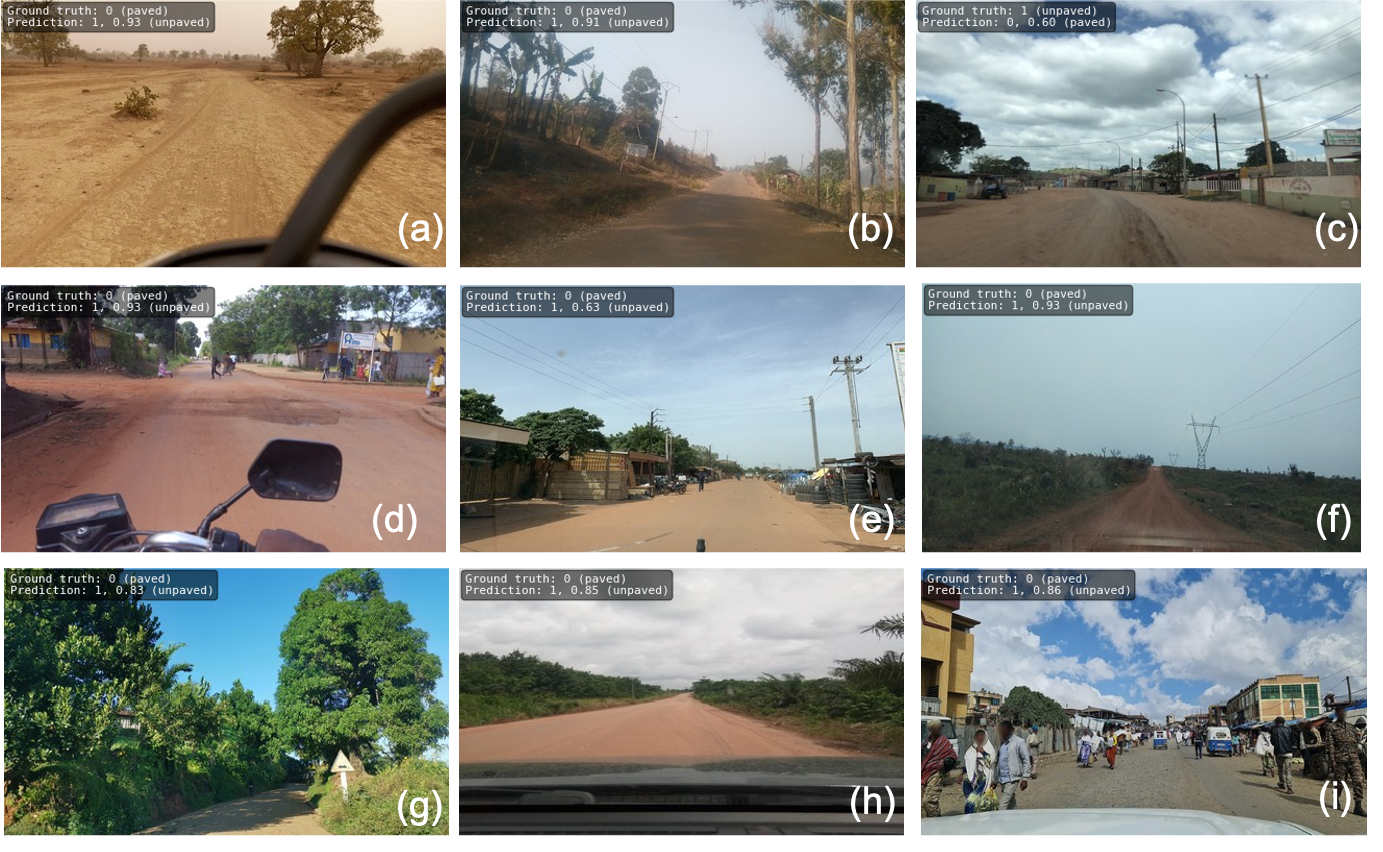}
  \caption{Examples of False Positives and False Negatives for the SWIN Transformer based classification model for road surface prediction on Mapillary based street-view images. Ground Truth, Prediction and confidence score are indicated in the top left corner of each image.
  }
  \label{fps}
\end{figure}

\begin{figure*}[htbp]
  \centering
  \includegraphics[width=1.0\linewidth]{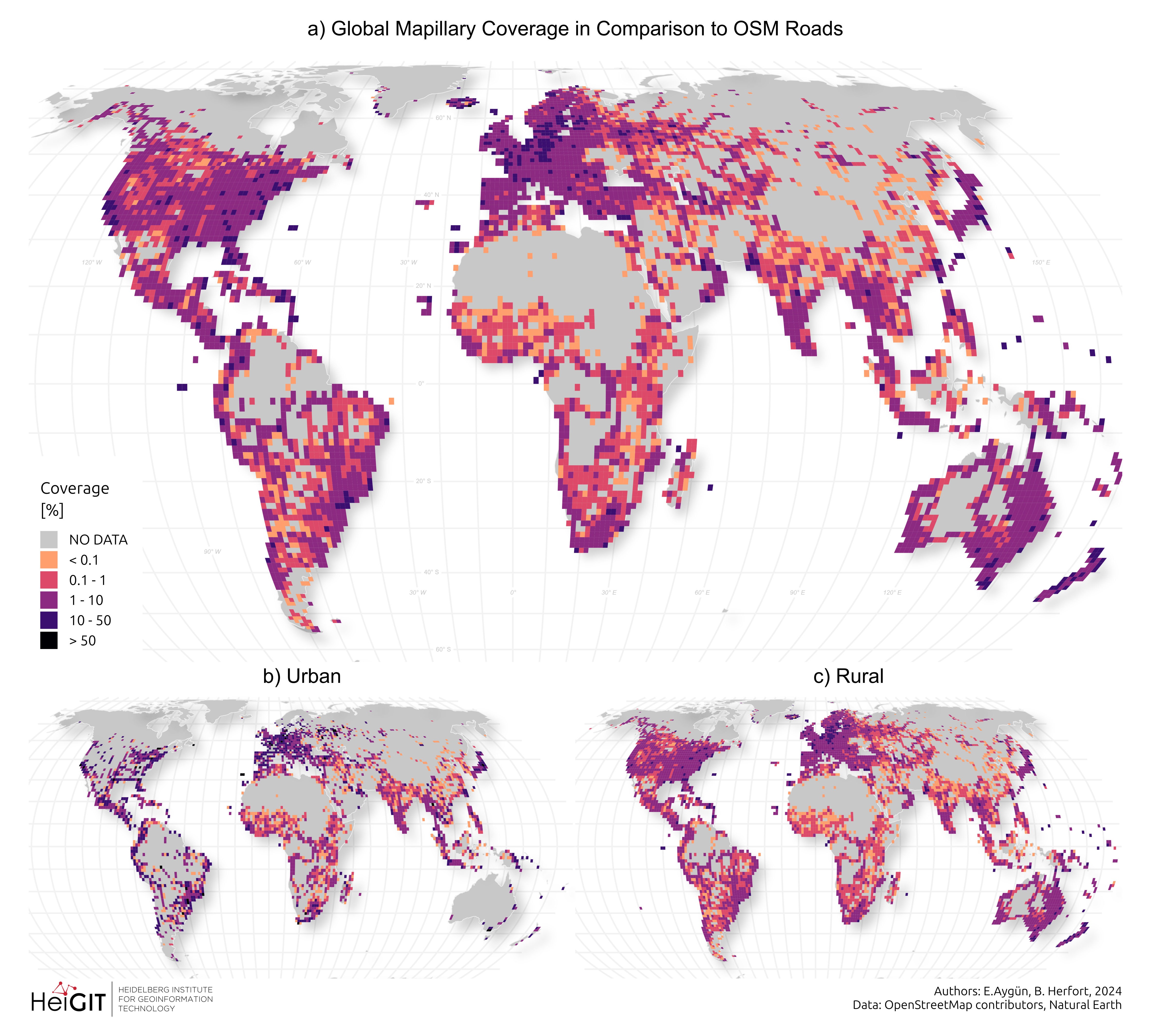}
  \caption{Global overview of spatial distribution of OpenStreetMap (OSM) road length coverage derived from Mapillary data (based on zoom 8 tiles). The maps illustrate: (a) Total OSM Road Length Coverage, (b) Urban OSM Road Length Coverage, and (c) Rural OSM road Length Coverage, showing the varying degrees of coverage across different regions of the world.  The color gradient from blue to yellow indicates the percentage of length coverage, with lighter colors representing higher coverage. }
  \label{mapillarycoveragemap}
\end{figure*}

To further analyze its performance, we reviewed True Positives and False Positives (Fig. \ref{tps} and Fig. \ref{fps}, respectively), highlighting instances where the model succeeded or struggled to detected paved road surface.
The model handled diverse scenarios, including occlusions caused by cars or animals, quite well.
However, challenges arose in situations where road surfaces were covered with mud, gravel or water (Fig. \ref{tps} (b),(c),(i) or (d)).
These instances occasionally reduced the model's confidence or resulted in incorrect predictions, reflecting the inherent difficulty of these mixed surface types even for human annotators.

Table \ref{tab:osm_model_performance} compares the results of our deep learning model against OSM data for different global regions.
The model achieved the F-1 scores for paved roads, ranging from 91 to 97 \%, which indicated a very good model performance across all global regions.
Since this information is aggregated on continent level, one could expect the class imbalance to be skewed (with paved roads being the dominant target class).
The model tended to overestimate unpaved roads, as indicated by the relative high number of False Negatives.
In fact, for all continents except Oceania the number of False Negatives was higher that the number of True Negatives.

Hence, we considered Matthews correlation coefficient (MCC) for a more nuanced evaluation.
The model achieved MCC values from 0.288\% to 0.52\%. 

\begin{table*}[h]
\centering
\caption{Comparison of OSM road surface information and predictions by continent.}
\resizebox{\textwidth}{!}{%
\begin{tabular}{@{}lrrrrrrrrrr@{}}
    \toprule
    Continent      & Total Images & True Positives  & False Positives & True Negatives & False Negatives & Accuracy   & F1-Score & Precision & Recall & MCC \\ \midrule
    Africa         & 181,486   & 139,329   & 8,452    & 15,114   & 18,591   & 0.851 & 0.912 & 0.943    & 0.882 & 0.453 \\
    Asia           & 923,597   & 844,712   & 17,760   & 14,229   & 46,896   & 0.930 & 0.963 & 0.979    & 0.947 & 0.288 \\
    Europe         & 3,173,008  & 2,870,510  & 97,626   & 96,586   & 108,286  & 0.935 & 0.965 & 0.967    & 0.964 & 0.45 \\
    North America  & 1,352,035  & 1,264,493  & 26,606   & 17,177   & 43,759   & 0.948 & 0.973 & 0.979    & 0.967 & 0.306 \\
    Oceania        & 268,061   & 243,414   & 8,082    & 9,475    & 7,090    & 0.943 & 0.970 & 0.968    & 0.972 & 0.525  \\
    South America   & 557,462   & 477,113   & 15,938   & 20,805   & 43,606   & 0.893 & 0.941 & 0.968    & 0.916 & 0.387 \\
    \bottomrule
\end{tabular}
}
\label{tab:osm_model_performance}
\end{table*}

\subsection{Mapillary Coverage Analysis}

Before analysing global spatial distribution of road pavedness, we need to evaluate the coverage of the Mapillary dataset.
In general, Mapillary's global coverage was spatially very uneven (cf. Figure \ref{osm_surface_coverage_composite3} (a)). 
Most western European countries, North America and Australia were better covered than other parts of the world.
For Asia, Africa and South America, many countries showed a low coverage. Russia and China showed strong heterogeneity in terms of Mapillary data coverage.
Japan, Thailand and some parts of India showed good coverage while Pakistan, Iran and Indonesia showed notably sparse coverage.
The average Mapillary global coverage was 3.48\% with hot spots of about 70\% coverage in cities such as Antwerp, Lille, Würzburg, Leipzig, Berlin, Moscow, Detroit, Boston, Washington D.C, Melbourne.

We observed a notable distinction in the spatial distribution of Mapillary coverage for urban and rural areas (cf. Fig. \ref{osm_surface_coverage_composite3} (b) and (c)).
On an average urban areas showed higher coverage (8.88\%) than rural areas (2.65\%).
Australia, Europe and North America showed only marginal coverage differences between urban and rural areas.
In contrast, the divide was most pronounced for South America, Africa and parts of Asia (cf. Fig. 14 in Appendix).

Nevertheless, our model results add surface information for 3.92 Million kilometers of roads for places where this information was previously unavailable (cf. Table \ref{tab:mapillary_coverage_per_continent}).
In total our combined road surface dataset sourced from OSM and Mapillary covers a length of 36.7 Million kilometers of roads.
Globally, this dataset increases the coverage of surface information from 33\% (OSM only) to about 36\%.
A particular high increase could be observed for North America, where coverage increased by 6 percentage points.

\begin{table*}[h]
\centering
\caption{Coverage of road surface information from Mapillary and OSM per continent.}
\begin{tabular}{@{}|l|r|rr|rr|rr|@{}}
    \toprule
                   & OSM (total)  & \multicolumn{2}{|c|}{OSM (surface)} & \multicolumn{2}{|c|}{Mapillary} & \multicolumn{2}{|c|}{Combined}           \\ 
    Continent      & [$10^{6} km$]          & [$10^{6} km$]  & [\%]        & [$10^{6} km$] & [\%]  & [$10^{6} km$]    & [\%] \\
    \midrule
    Africa         & 14.11              & 5.93       & 42\%   & 0.66    & 4\%    & 6.27  & 44\%  \\
    Asia           & 32.53              & 6.20       & 19\%   & 1.86    & 6\%    & 7.29  & 22\%  \\
    Europe         & 21.07              & 7.72       & 37\%   & 2.27    & 11\%   & 8.45  & 40\%  \\
    North America  & 21.70              & 6.65       & 31\%   & 2.51    & 12\%    & 8.08  & 37\%  \\
    Oceania        & 2.30               & 1.58       & 69\%   & 0.43    & 19\%    & 1.70  & 74\%  \\
    South America  & 8.89               & 4.71       & 53\%   & 0.64    &  7\%    & 4.91  & 55\%  \\ 
    \midrule
    Total          & 100.61             & 32.78      & 33\%   & 8.38    &  8\%    & 36.70 & 36\%  \\ 
    \bottomrule
\end{tabular}
\label{tab:mapillary_coverage_per_continent}
\end{table*}

\begin{figure*}[h]
  \centering
  \includegraphics[width=1.0\linewidth]{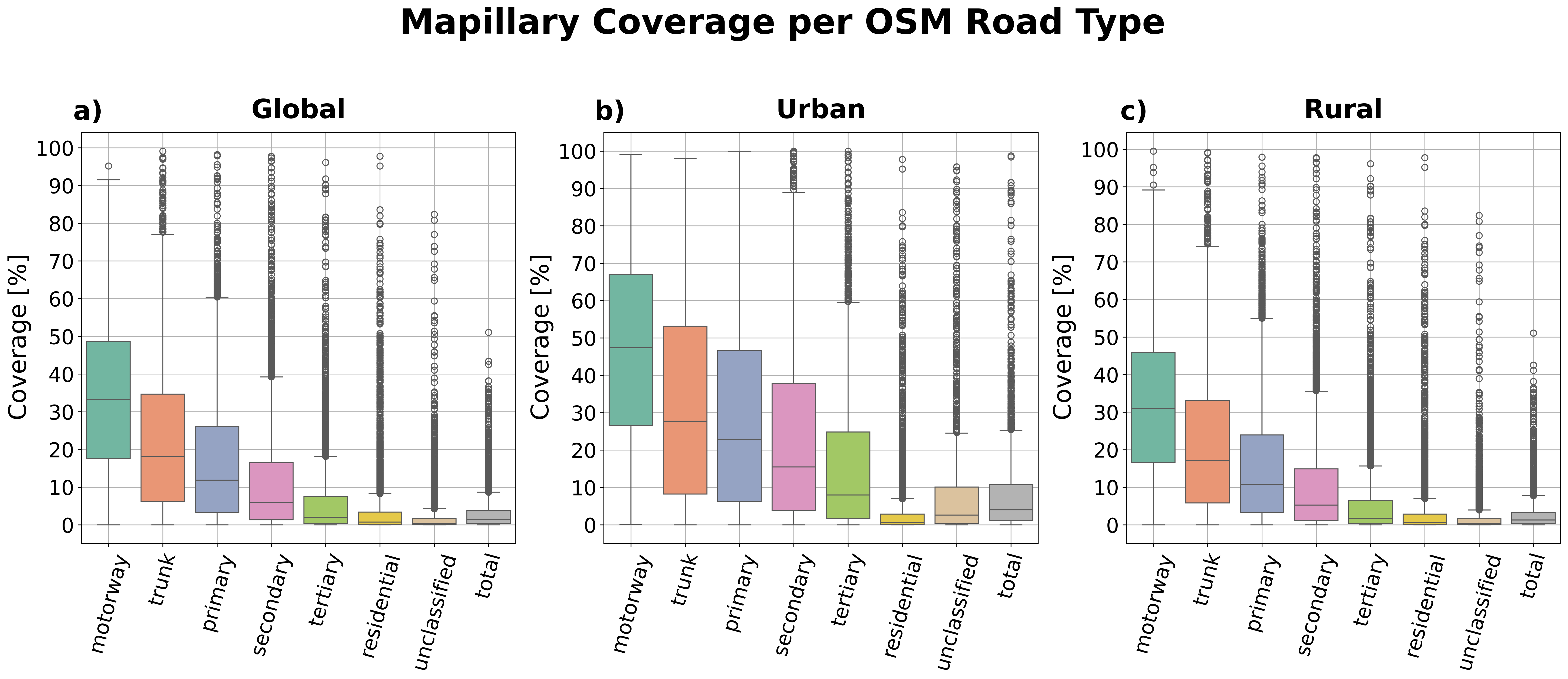}
  \caption{ Percentage of road length covered by Mapillary for different OSM road classes (as indicated by the highway tag) across Continents; (a) shows total Mapillary length coverage per continent whereas (b) and (c) depict the same for urban and rural areas for each continent respectively. OSM segment type importance decreasing from left to right along x-axis.}
  \label{coverageperosmroad}
\end{figure*}

Mapillary coverage varied also considering different OSM road classes (cf. Fig. \ref{coverageperosmroad} (a)).
The most important roads in a country's system such as motorways and trunks showed a higher global Mapillary coverage of about 45\% with a steady decrease in the coverage observed for less important roads.
Residential and unclassified streets showed a sharply staggering coverage ranging from 2 - 5\%. 
A similar pattern was observed for urban and rural areas with 65\% and 40\% coverage for most important roads respectively (Figure \ref{coverageperosmroad} (b) and (c)). 
The presence of outliers and the spread of the data also indicate inconsistencies in coverage for less dominant road types, such as foot ways or unclassified roads, across different regions.
  
\subsection{Global Road Surface Analysis}
\begin{figure*}[htbp]
  \centering
  \includegraphics[width=1.0\linewidth]{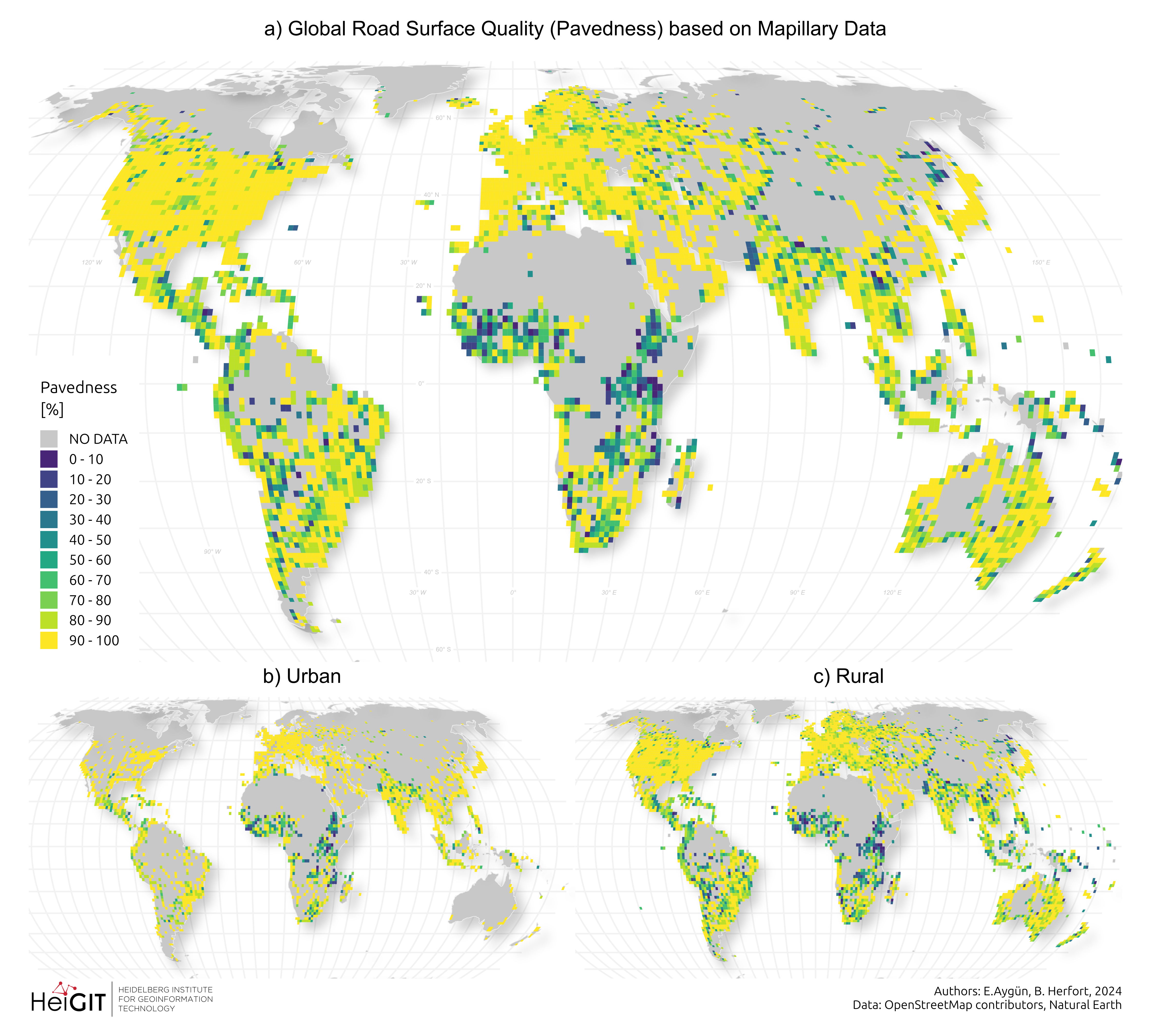}
  \caption{Distribution of road surface quality predicted based on Mapillary data: (a) total pavedness (defined as the ratio of total paved roads w.r.t total OSM roads for each zoom 8 tile), (b) and (c) pavedness per tile, calculated for urban and rural areas respectively.}
  \label{osm_surface_coverage_composite3}
\end{figure*}

After compiling the global dataset using Mapillary street-view imagery, we visualized the aggregated results on zoom 8 tiles to better understand global road surface patterns.
These patterns for paved ratio (or state of infrastructure) are displayed in Fig. \ref{osm_surface_coverage_composite3}.
Distinct areas with a proportion of paved roads (colored in yellow or orange) less than 40\% include countries like Pakistan, Nepal, Rwanda, Burundi, parts of Tanzania, Sierra Leone, Guinea, Laos, Mozambique, Mali, and Bolivia.
In contrast, large regions show medium to high paved ratios (between 60\% and 80\%) — notably across South America, Asia, Eastern Europe, Mexico, India, and Australia.
Table \ref{tab:paved-coverage} provides a breakdown of these results per continent, as well as for urban and rural areas.

\begin{table*}[h]
\centering
\caption{Paved ratio by continent and country. Lowest 15 countries are displayed ordered by pavedness.}
\begin{tabular}{@{}|l|r|r|r|@{}}
    \toprule 
    Region         & Total    & Urban   & Rural \\
    \midrule
    Africa         & 0.768    & 0.784      & 0.707     \\
    Asia           & 0.917    & 0.930      & 0.871     \\
    Europe         & 0.934    & 0.959      & 0.911     \\
    North America  & 0.951    & 0.957      & 0.938     \\
    Oceania        & 0.941    & 0.970      & 0.888     \\
    South America  & 0.860    & 0.874      & 0.804     \\ 
    \midrule
    Sierra Leone   & 0.213 	  & 0.210      & 0.221       \\
    Gambia 	       & 0.359    & 0.304      & 0.721      \\
    Kenya 	       & 0.364 	  & 0.421      & 0.224       \\
    DR Congo       & 0.451 	  & 0.536      & 0.288       \\
    Uganda 	       & 0.523    & 0.551      & 0.434       \\
    Pakistan 	   & 0.527 	  & 0.855      & 0.285       \\
    Laos           & 0.539    & 0.932      & 0.391 	   \\
    Zambia 	       & 0.592    & 0.678      & 0.442       \\
    Timor-Leste    & 0.595 	  & 0.624      & 0.561       \\
    Nigeria 	   & 0.628 	  & 0.649      & 0.395       \\ 	
    Togo 	       & 0.635 	  & 0.637      & 0.610       \\
    Haiti 	       & 0.649    &	0.693      & 0.519       \\
    Mayotte 	   & 0.659 	  & 0.696      & 0.631 \\
    Mali 	       & 0.660 	  & 0.759      & 0.332 \\
    Tanzania       & 0.669    & 0.667      & 0.679 \\
    \bottomrule
\end{tabular}
\label{tab:paved-coverage}
\end{table*}

\begin{figure*}[h]
  \centering
  \includegraphics[width=1.0\linewidth]{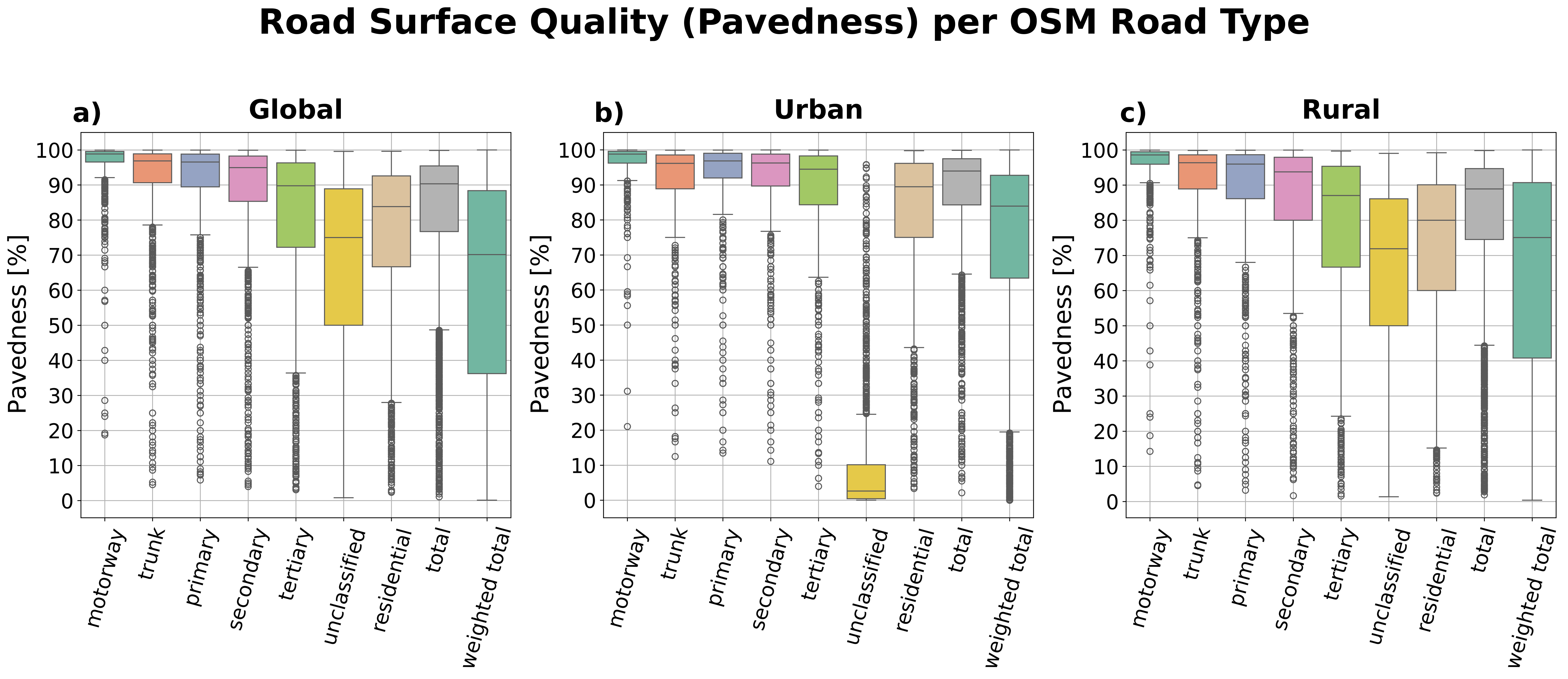}
  \caption{Distribution of Surface Pavedness across continents by OSM road type, visualizing how road surfaces (paved or unpaved) are distributed globally and regionally for various OSM highway tags. The data is divided into three main categories: Total pavedness (a), urban pavedness(b), and rural pavedness (c).}
  \label{pavedperosmroad}
\end{figure*}

Motorways and primary roads are mostly paved across all continents and in both urban and rural areas as seen in Figure \ref{pavedperosmroad}.
Residential roads and tertiary roads show the most variation in pavedness, particularly in rural areas. 

\begin{figure*}[h]
  \centering
  \includegraphics[width=1.0\linewidth]{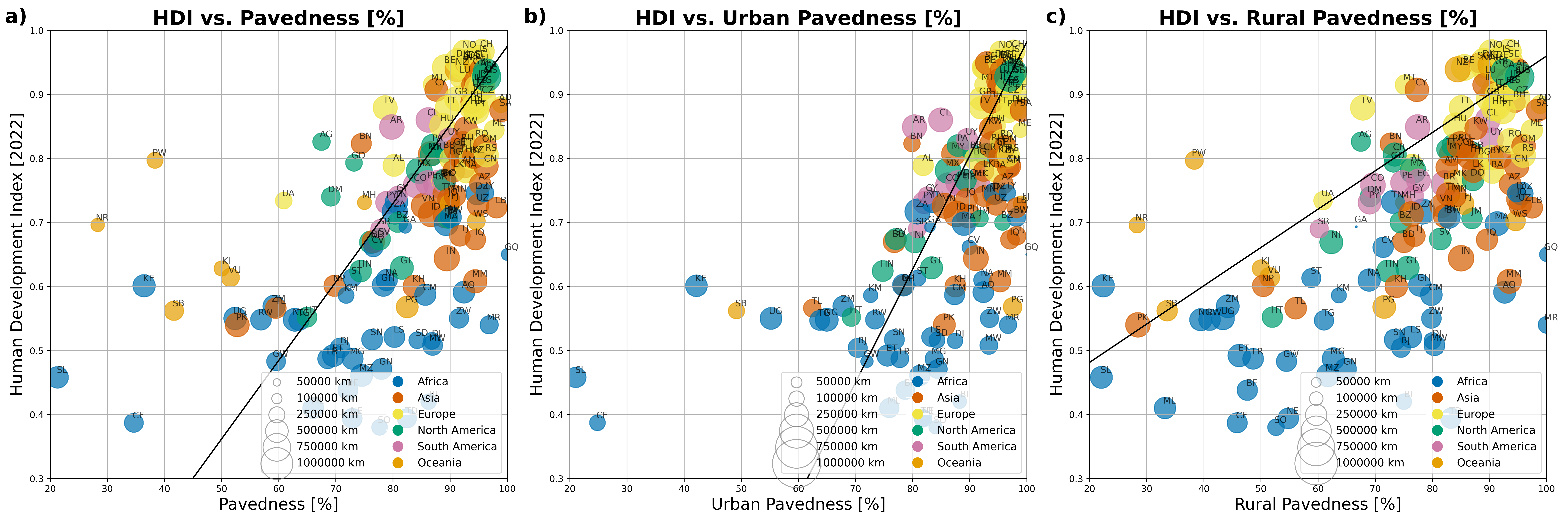}
  \caption{Human Development Index (HDI) against pavedness(\%) - i.e paved road infrastructure in different countries across continents. The size of the circles represents the total road length, indicating the extent of road infrastructure as mapped in OSM. There is a positive correlation, with higher HDI values associated with countries having a higher percentage of paved roads. The circles in (a) depicts HDI w.r.t total pavedness for the entire country, whereas (b) and (c) show the same for urban and rural areas of the country respectively. $\texttt{coeff}_{\texttt{pearson}}$ for the total, urban and rural cases are (0.54, 0.61, 0.62) respectively while the corresponding $\textbf{R}^2$ values are (0.29, 0.37, 0.39).}
  \label{hdipavedratio}
\end{figure*}

A comparison of the pavedness metric against the HDI (cf. Fig \ref{hdipavedratio}(a)) indicated a positive correlation between HDI and total pavedness at the country scale.
Europe and North America dominated the upper right corner with high HDI and high road pavedness, whereas Africa and Oceania trail towards the bottom left.
Oceania island countries were high on HDI but low on pavedness.
A similar pattern was observed for the association between HDI and both urban and rural pavedness (cf. Fig \ref{hdipavedratio}(b) and (c)):
Europe and Asia exhibited higher rural pavedness, also contributing to their elevated HDI, while African countries clustered at the lower end of both metrics. 
However, it is noteworthy to mention that global and rural results (Fig \ref{hdipavedratio}(a) and (c) were more dispersed in comparison to urban results (Fig \ref{hdipavedratio}(b)) that tended to concentrate.
This was likely due to the variation of HDI which is more distinct in rural areas across countries than for urban areas. 

\section{Discussion}
In this study we propose a method to enrich OSM road geometry data with road surface type information utilizing Mapillary street level imagery.
The deep learning model identified paved roads with a very high F-1 score of about 95\% and overall accuracy of 92\%.
In comparison, the StreetSurfaceVis model also utilizes street-level imagery and achieves a similar test data accuracy of 91\%, however can also be considered less challenging, as they use only about 9,000 images primarily located in one country \citep{kapp_streetsurfacevis_2024}.
Using inertial sensor data represented by accelerometers and gyroscopes, \cite{menegazzo_road_2021} report a validation accuracy of 93\% for CNN-based deep neural network road surface type classification.
\cite{zhou_mapping_2024} utilize Google satellite imagery and a convolutional neural network to classify road surfaces into two classes (paved and unpaved) in Kenya and achieve an accuracy of 94\%.
As such, our approach achieved comparable performance to existing approaches, while extending coverage to the global scale and very heterogeneous geographies and utilizing only openly available data.

We applied the road surface classification model on a global scale, processed about 105 Million Mapillary images and derived a combined road surface dataset from OSM and Mapillary.
This dataset increases the coverage of openly available global road surface information by more than 3 Million kilometers covering now about 36\% of the entire global road network length.
Our results emphasized the uneven global coverage of Mapillary data across continents, between urban and rural areas and for various road types shown by other authors \citep{quinn_every_2019, mahabir_crowdsourcing_2020}.
Furthermore, our global road surface analysis results revealed disparities in infrastructure quality across continents at a high spatial resolution.Figure 15 in Appendix presents a comprehensive summary map illustrating the global road surface analysis at the country level.

However, our analysis also comes with unavoidable limitations that need to be considered when utilizing our dataset for global scale analysis.
These limitations can be attributed towards three major factors: model uncertainties, spatial bias in Mapillary and OSM data coverage, and bias in road type representation.

The SWIN-transformer based deep learning model was error prone for street view images that do not show roads.
In fact, it has been shown by \cite{biljecki_street_2021} that the inevitably highly heterogeneous quality of Mapillary images poses challenges for many research applications, especially if computer vision techniques are used.
In our case, we had to employ another two deep learning model to filter out these images which introduces further sources of uncertainty.
As such, surface type classification model performance thus also indirectly depends on our ability to detect and remove non-road images.
Whereas model performance was good overall, still our model overestimated the presence of unpaved roads as reflected by the relative high number of False Negatives in comparison to the overall number of non-paved classification. The misclassification was primarily caused by the heterogeneous quality of Mapillary data.
A significant portion of the incorrectly classified images lacked clear, direct street-view imagery. Instead, many of these images captured side views of houses, driveways, or landscapes, rather than offering frontal views of roads or streets (cf. Figure 16 in Appendix).
\cite{hou_comprehensive_2022} present a comprehensive framework to assess the quality of street view imagery (e.g. factors among others: blurriness, illumination conditions, obstruction, distortion), which should be used in future work to improve selection of "proper" images.

Visual inspection revealed that a smaller fraction of mislabeling in the False Positives and False Negatives dataset was attributed to human error (uncertainties in OSM data quality) rather than model performance, however it was not possible to quantify this effect on a larger scale.
This uncertainty became prevalent when distinguishing between road types that visually blend together.
Another modelling uncertainty was introduced by map-matching between OSM road geometries and Mapillary image locations, which can lead to wrong attribution in situations where a Mapillary image is located closer than 10 meters to more than one road.
As a consequence, the same Mapillary image might be attributed towards several roads and estimated Mapillary coverage might be too high.
\cite{newson_hidden_2009} describe how Hidden Markov Models (HMM) could be used to improve our simplistic map matching by accounting for GPS measurement noise and the layout of the OSM road network.

Spatial bias becomes evident when considering the uneven geographic distribution of Mapillary data.
We could only apply our road surface classification model in places where Mapillary images have been captured.
Many authors have studied spatial bias in Mapillary coverage and conclude that limited spatial coverage is still a major factor that prevents wider adaptation in urban studies \citep{biljecki_street_2021, mahabir_crowdsourcing_2020}. 
In respect to our study, our results might not be representative for all roads in regions where Mapillary coverage was very low, and thus led to biased pavedness estimates.
Furthermore, OSM road coverage itself is not 100\% in all places and contains an inherent spatial bias as well \citep{worldusermap}. 
Hence, our estimated total road length might be too small in countries with lower OSM data quality, e.g. due to lower number of contributors.
In particular, this will have an effect on results for high-income countries (China, India, Brazil, etc.) where it is known that the OSM data gap is still big \citep{herfort_spatio-temporal_2023}.
Together these factors resulted in overestimated Mapillary coverage in places with uncertain OSM quality, e.g. in rural areas.
It is also possible that tiles with lower Mapillary coverage (i.e., fewer than 100 OSM segments covered) may not accurately reflect the characteristics of the entire tile. 

Beyond spatial bias there is a bias in road types represented in our dataset.
In particular for our global pavedness estimates, residential and tertiary roads are underrepresented.
As shown Mapillary coverage is skewed towards road types which are more likely to be paved (e.g. motorways or trunks).
As a consequence our estimated pavedness values will be too high.
This should be considered in global analysis, e.g. when deriving pavedness metrics on the country scale.
We see this reflected by the lower estimate for road pavedness of 53\% of \cite{african2014tracking} in comparisons to the pavedness we have estimated for Africa (77\%).

With the help of big data, machine learning, and geospatial analysis, this dataset has been meticulously curated to offer valuable insights for advancing research and supporting data-driven decision-making.
Based on our results we propose the following recommendations: 

\begin{itemize}
       \item The value of this dataset is evident as a critical tool for various socio-economic applications, offering crucial insights for infrastructure planning, balancing development with environmental conservation, and monitoring emissions. It also plays a significant role as an economic development indicator, providing essential data for evaluating several Sustainable Development Goals (SDGs), which is directly linked to economic growth, access to services, and overall connectivity in both urban and rural areas.
       \item This release of this dataset has several key benefits for the scientific and research community, as well as for the broader public. Besides, we hope that this work enables further innovation, collaboration and research and could be used as a benchmark for testing new algorithms, models, or methodologies.
       \item Mapillary's data coverage gap needs to be filled by building local capacity at the city level by empowering local stakeholders, municipalities and local NGOs to systematically collect street-level imagery.
       Furthermore, communities should have access to openly available and free of cost street-level imagery for various tasks beyond just mapping road surfaces, facilitating a wider range of geospatial analysis and decision-making processes.
\end{itemize}


As part of future work, we would like to move a step further to address the still remaining data gaps on road surface information by utilizing a combination of OSM data, Mapillary street level images and satellite imagery. Furthermore, we aim to develop a deep-learning based tool capable of distinguishing between proper frontal street-view imagery and side-view imagery. This enhancement would greatly improve the quality of datasets sourced from Mapillary by filtering out non-relevant images, such as those of landscapes or buildings and positively impact various derived applications. 
The work could furthermore be extended to extract other valuable road attribute information, e.g. related to bikeability and walkability.
One potential challenge in enhancing data usability is facilitating its integration into OpenStreetMap (OSM), which could significantly enhance OSM's data quality. Leveraging AI-driven strategies via AI-based editors to assist in this process  would help streamline human contributions to mapping, thereby accelerating the mapping effort. This could play an important role in the future as well by aiding in efficient and scalable updates to OSM.
\section*{Dataset Availability}
The dataset generated and analyzed during the current study is openly available in the [Repository Name] at [DOI or URL].
We have enriched the dataset with a set of attributes with relevant Mapillary as well OSM segment information (corresponding to the OSM road segment matched to the Mapillary image point) for each image point, thereby facilitating further usage of this dataset for a variety of downstream analyses in a context of geospatial applications or computer vision modelling benchmarking efforts.
\section*{Disclaimer}
Last but not the least, we do not want readers to develop a bias towards paved roads, as both paved and unpaved roads present unique advantages and challenges. It is true that in many developing countries, the presence of paved roads directly affects access to essential services like healthcare, education, and markets and that paved roads are often critical for poverty reduction by linking isolated communities to economic opportunities. While paved roads offer durability and ease of travel in high-traffic areas, unpaved roads remain important for their cost-effectiveness, adaptability to local conditions, and role in connecting remote regions. In Africa, where infrastructure is still developing in many areas, unpaved roads are a vital link for rural economies, communities, and ecosystems. The environmental benefits of unpaved roads cannot be understated given that unpaved roads can be constructed with minimal disruption to the environment, preserving local ecosystems. Unpaved roads support sustainable construction and allow water to permeate into the ground, reducing runoff and helping with water absorption into the soil. Though they reduce environmental impacts, if not properly managed, can lead to erosion, loss of soil quality, and exacerbate environmental degradation \citep{Yang2024}\citep{ESA2021}. This is especially important in sensitive ecological zones or areas prone to heavy rainfall, where unpaved roads can contribute to landslides or flooding\citep{ESA2021}.

\section*{Acknowledgements}
 We would like to express our sincere thanks to the HeiGIT CrowdMap web application team for their support in facilitating the creation of our labeled dataset.
 Special gratitude goes to the participants of the July 2023 mapathon at Heidelberg University, whose invaluable contributions helped classify a large number of Mapillary images. The authors would like to thank Dr. Michael Auer for his support on big data analytics and Clemens Langer for his initial involvement and assistance in data analysis and mapping.
 The authors gratefully acknowledge support for the High-Performance Computing Infrastructure by the state of Baden-Württemberg through bwHPC and the German Research Foundation (DFG) through grant INST 35/1597-1 FUGG.
 Moreover, the authors acknowledge the data storage service SDS@hd supported by the Ministry of Science, Research and the Arts Baden-Württemberg (MWK) and the German Research Foundation (DFG) through grant INST 35/1503-1 FUGG. 
 We would like to extend our heartfelt thanks to the Klaus-Tschira Stiftung for their financial support.

\bibliographystyle{elsarticle-harv} 
\bibliography{main}
\section*{Appendix}
This is the appendix.

\begin{figure*}[h]
    \centering
    \includegraphics[width=0.6\linewidth]{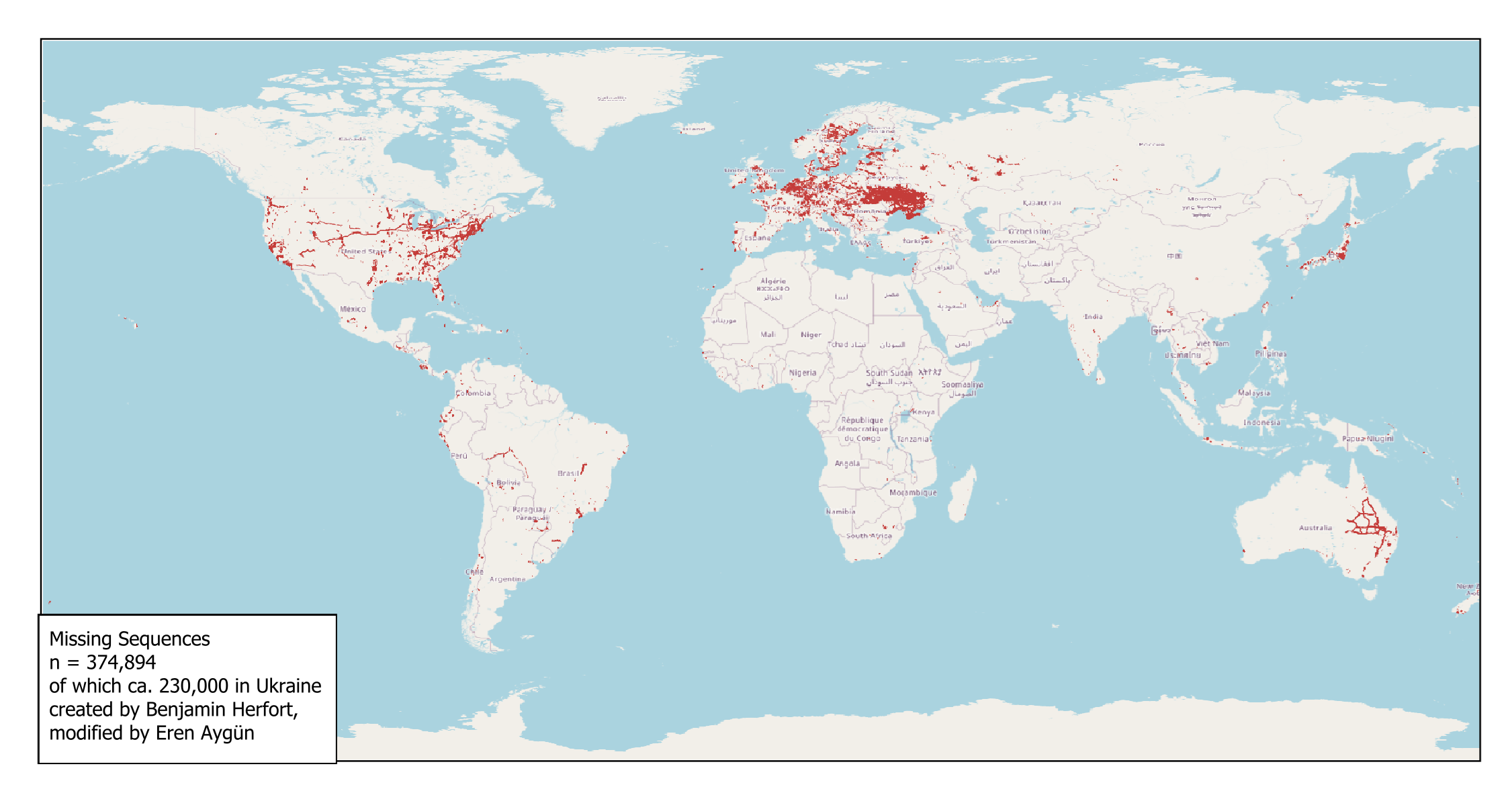}
    \caption{Distribution of sequences that could not be downloaded despite multiple attempts. Basemap: \textcopyright OpenStreetMap contributors.}
    \label{fig:16}
\end{figure*}

\begin{table*}[h!]
\label{tab:attributes}
\begin{adjustbox}{max width=\textwidth}
\centering
\begin{threeparttable}
\caption{Description of attributes in the final dataset. This consists of Mapillary image metadata, information derived during the deep learning procedures, information about the OSM object the image was matched to as well as attributes used to join information with auxillary data.}
\begin{tabular}{cc}
\hline
\textbf{Attribute} & \textbf{Description} \\
\hline
\multicolumn{2}{c}{\textbf{Mapillary related attributes}}\\ 
\hline
\texttt{sequence} & The unique identifier for the sequence of images to which this image belongs. Originally \textit{sequence\_id}. \\
\texttt{id} & The unique identifier for the image within the Mapillary database. \\
\texttt{url} & The direct link to access the image data.Originally \textit{thumb\_original\_url}. \\
\texttt{geometry} & location of image as Point geometry of after Mapillary's image processing. Originally \textit{computed\_geometry}. \\
\texttt{long}, \texttt{lat} & The longitude and latitude values representing the geographic location where the image was captured, extracted from \textit{computed\_geometry} \\
\texttt{height}, \texttt{width} & The height and width of the original image in pixels. \\
\texttt{altitude} & The altitude (in meters) above sea level at which the image was taken. Originally \textit{computed\_altitude}.  \\
\texttt{make} & The brand of the camera used to capture the image. \\
\texttt{model} & The model of the camera used to capture the image. \\
\texttt{creator} & Information about the user who took the image, including the user's ID and username. \\
\texttt{is\_pano} & A boolean value indicating whether the image is panoramic (1 for true, 0 for false). \\
\texttt{timestamp} & The date and time when the image was captured. Originally \textit{captured\_at}.\\
& \\
\hline
\multicolumn{2}{c}{\textbf{DL related attributes}} \\
\hline
\texttt{pred\_label} & Prediction label for the surface type (1 for unpaved, 0 for paved). \\
\texttt{pred\_class} & The classification output of the prediction model. \\
\texttt{pred\_score} & The confidence score of the prediction model. \\
\texttt{zs\_pred\_class} & The zero-shot prediction class estimated by the CLIP model. \\
\texttt{zs\_pred\_score} & The z-test score of the prediction. \\
\texttt{road\_pixel\_percentage} & The ratio of the number of pixels estimated to be road to the total number of pixels in the image. \\
\texttt{no\_road\_image\_filter} & A boolean filter indicating whether the image contains a road (1 for yes, 0 for no). \\
& \\
\hline
\multicolumn{2}{c}{\textbf{Attributes related with the OSM object the image has been matched to}} \\
\hline
\texttt{count} & The number of OpenStreetMap (OSM) segments to which the image point has been assigned. \\
\texttt{osm\_ids} & The OSM IDs of the OSM segments to which the image point has been assigned. \\
\texttt{osm\_tags\_highways} & The highway tags associated with the assigned OSM segments. \\
\texttt{osm\_tags\_surfaces} & The surface tags associated with the assigned OSM segments. \\
\texttt{distances\_meter} & The distances (in meters) from the image point to the respective OSM segments. \\
\texttt{changeset\_timestamps} & The timestamp of the most recent update to the respective OSM segments. \\
\texttt{osm\_id} & The OSM ID of the primary OSM segment to which the image point has been assigned. \\
\texttt{osm\_tags\_highway} & The highway tag associated with the primary assigned OSM segment. \\
\texttt{osm\_tags\_surface} & The surface tag associated with the primary assigned OSM segment. \\
\texttt{distance\_meter} & The distance (in meters) from the image point to the primary assigned OSM segment. \\
\texttt{changeset\_timestamp} & The timestamp of the most recent update to the primary assigned OSM segment. \\
\texttt{abs\_dif} & The absolute distance difference between the assigned OSM segment and the nearest OSM segments.\\
\texttt{percent\_dif} & The percentage difference in distance between the assigned OSM segment and the nearest OSM segments.\\
\texttt{shortest line} & The LineString Geometry to the OSM segment the point has been assigned to in binary format\\
& \\
\hline
\multicolumn{2}{c}{\textbf{Geospatial Attributes}} \\
\hline
\texttt{continent} & The continent where the image was taken. \\
\texttt{country} & The country where the image was taken. \\
\texttt{longlat} & The longitude and latitude of the image represented as a list [longitude, latitude]. \\
\texttt{points\_tiles} & The x, y, z codes representing the tile on zoom level 8 where the image is located. \\
\texttt{z14\_tiles} & The x, y, z codes representing the tile on zoom level 14 where the image is located. \\
\texttt{ID\_HDC\_G0} & The Global Human Settlement Layer (GHSU) ID representing the urban area where the image was captured. \\
\texttt{UC\_NM\_LST} & The name of the urban area as per the GHSU Layer. \\
\texttt{agglosID} & The ID of the urban area according to the AFRICAPOLIS2020 dataset. \\
\texttt{agglosName} & The name of the urban area as per the AFRICAPOLIS2020 dataset. \\
\texttt{year\_and\_month} & The year and month when the image was captured based 
 on \textit{timestamp}. \\
\texttt{temporal\_index} & A monthly temporal index starting from January 2013 as 1; 0 is reserved for any date earlier than January 2013 based on \textit{timestamp}. \\
\hline
\end{tabular}
\end{threeparttable}
\end{adjustbox}
\end{table*}

\clearpage

\begin{figure*}[h]
  \centering
  \includegraphics[width=0.7\linewidth]{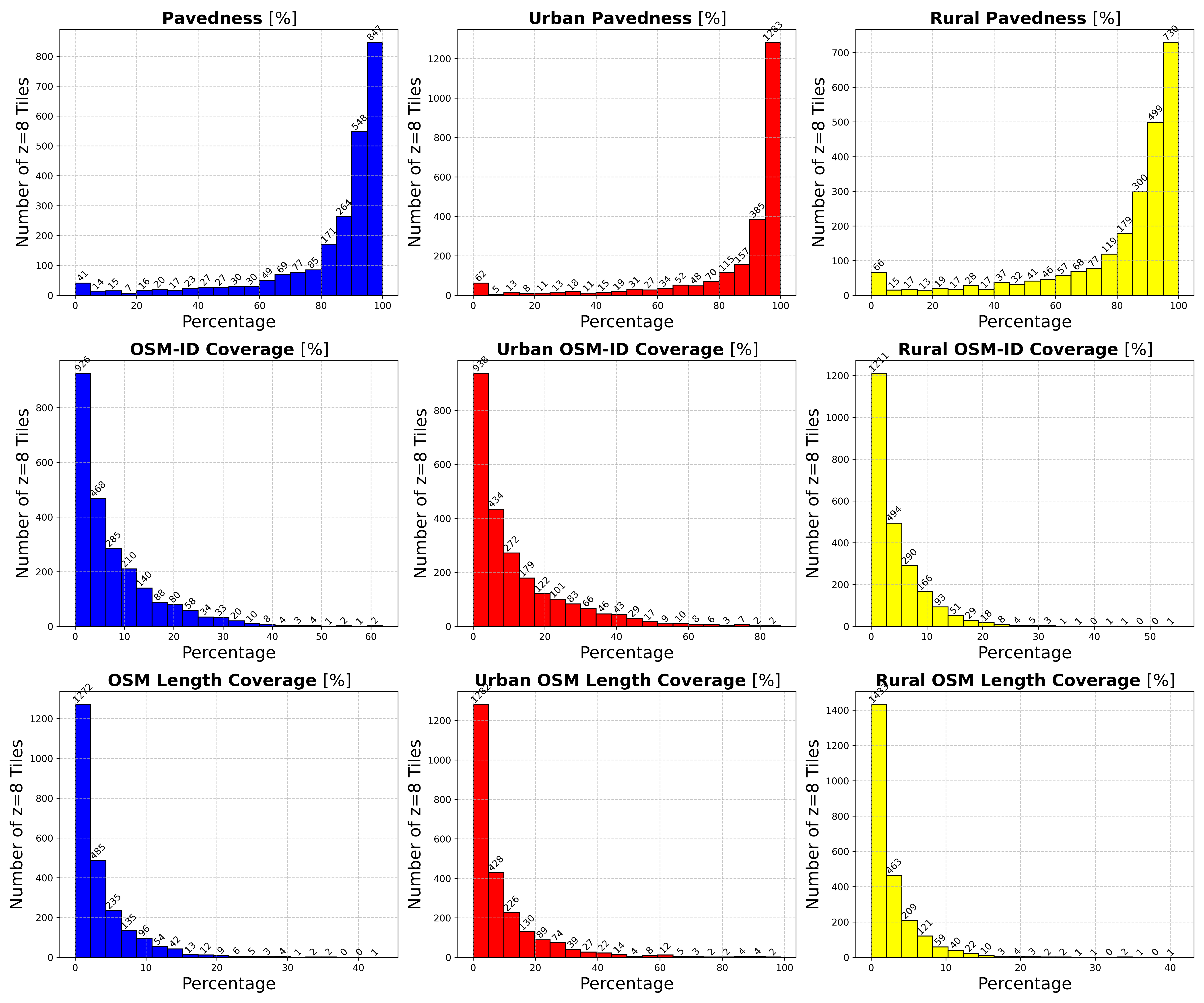}
  \caption{Distribution of predicted road surface pavedness and OpenStreetMap (OSM) road coverage metrics. (a) shows total pavedness (\%) whereas (b) and (c) show specific distributions across urban and rural regions respectively. (d)-(c) show total, urban and rural Mapillary coverage based on OSM road segment (indicated by OSM-ID)  while (g)-(i) shows Mapillary coverage in terms of aggregated OSM road length. Dashed orange lines represent the average values. }
  \label{histogram}
\end{figure*}

\begin{figure*}[htbp]
    \centering
    \includegraphics[width=\textwidth]{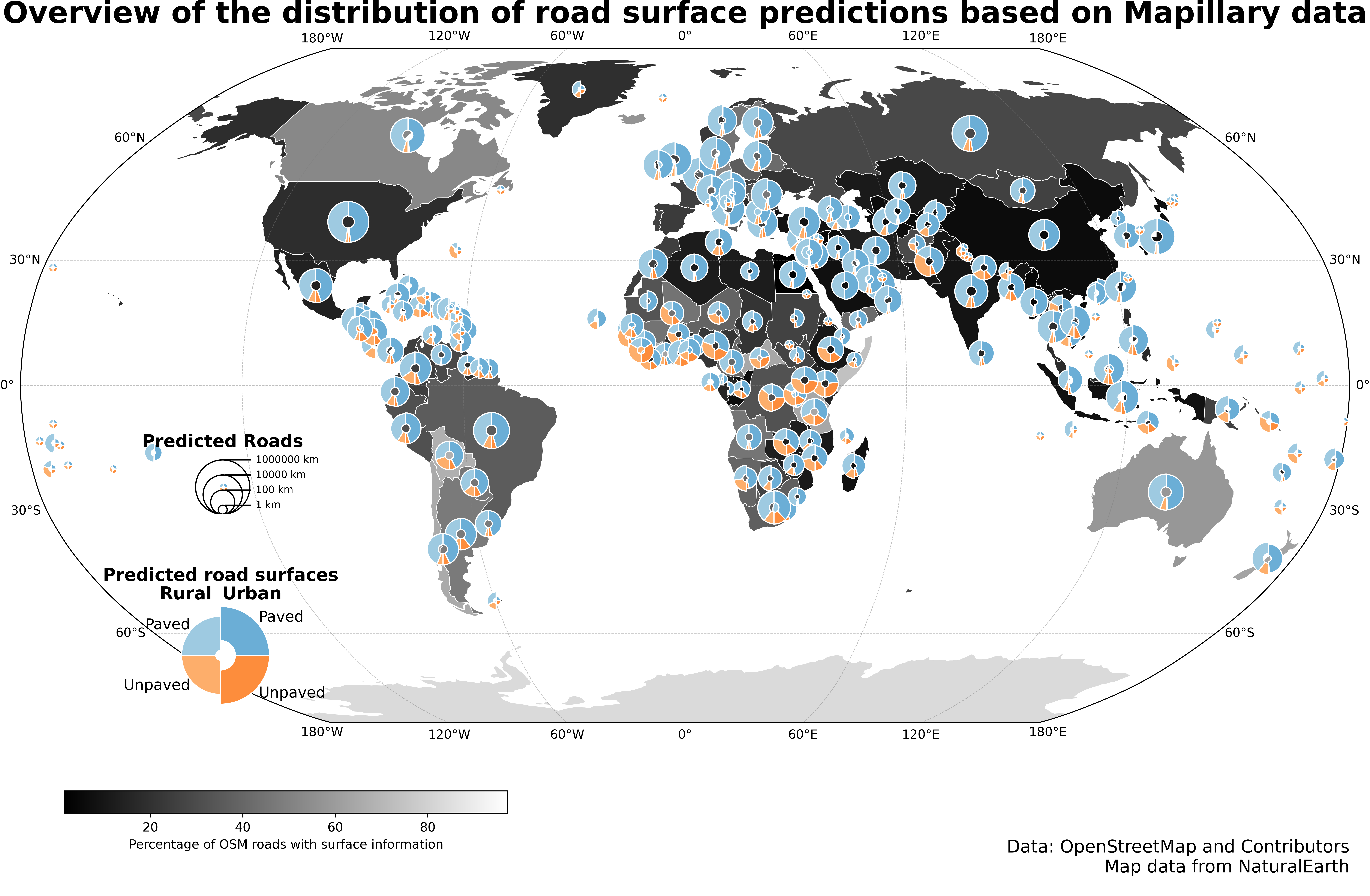}  
    \caption{Road surface predictions based on Mapillary data for rural and urban areas (left/right half of the pie chart respectively) along with the percentage of roads in OpenStreetMap(OSM) with surface information (choropleth map) at country level. Paved and unpaved road information is marked by blue and orange segments and the size of the semi-circle refers to the total length of predicted roads for that country, where a larger size indicates a more extensive road network.}
    \label{summaryresults}
\end{figure*}

Figure \ref{summaryresults} reveals patterns for both data coverage (Mapillary and OSM) as well as data extracted from our Deep learning Models on Mapillary Street-View imagery with the aim to fill the existing data gaps in OSM (notably extensive) in terms of road surface information.

As a summary of our results, Fig.\ref{summaryresults}, provides a country-level overview of road surface predictions(for rural and urban areas respectively) based on Mapillary data, showing the spatial global distribution of both paved and unpaved roads, along with the percentage of roads in OpenStreetMap (OSM) with surface information. Key Highlights are as follows:

\begin{itemize}
    \item The figure highlights regional differences in road surface infrastructure with developed regions tending to have a higher proportion of paved roads, not to mention better OSM road surface documentation.
    North America, Europe, and Australia have a large proportion of paved roads, as indicated by the dominant blue segments in their pie charts, especially in urban areas, whereas Africa, South America, and parts of Asia show more diversity in road surface types, with a significant presence of unpaved roads, especially in rural areas, marked by orange segments.
    \item The map uses a gradient from light gray to black to indicate the percentage of OSM roads with surface information. Lighter areas (e.g., parts of the US and Europe) have a higher percentage of roads with detailed surface data, while darker areas (e.g., much of Africa and Asia) have less surface information.

\end{itemize}

\clearpage

\begin{figure*}[h]
  \centering
  \includegraphics[width=0.7\linewidth]{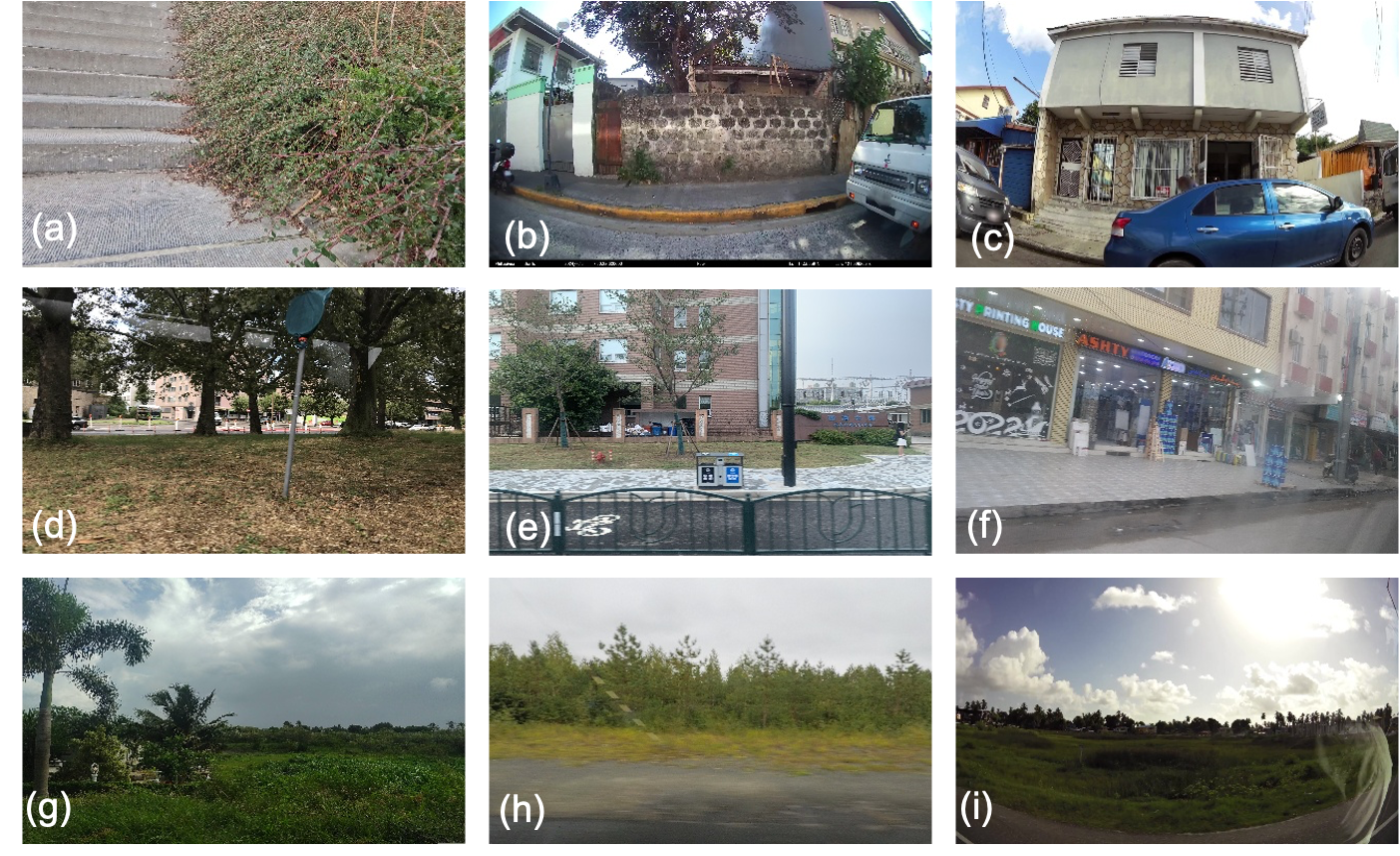}
  \caption{ Panels (a)-(i) demonstrate examples of poor-quality side-view images in Mapillary, where the primary focus is not on roads but rather on pavements, vegetation, houses, and various obstructions, such as walls and shops. These images lack a clear frontal view of the main roads, leading to challenges in classifying road surfaces or determining the quality of road infrastructure.}
  \label{AppendixFigsideviews}
\end{figure*}

\end{document}